\documentclass[10pt,twocolumn,letterpaper]{article}

\PassOptionsToPackage{numbers, sort&compress}{natbib}

\usepackage{iccv}   

\usepackage[utf8]{inputenc} %
\usepackage[T1]{fontenc}    %
\usepackage{url}            %
\usepackage{booktabs}       %
\usepackage{amsfonts}       %
\usepackage{nicefrac}       %
\usepackage{microtype}      %
\usepackage{xcolor}         %

\usepackage{color}
\usepackage{soul}
\usepackage{multirow}
\usepackage{xcolor}

\usepackage{wrapfig}
\newcommand{\cmark}{\ding{51}}%
\newcommand{\xmark}{\ding{55}}%

\definecolor{darkred}{rgb}{0.7,0.1,0.1}
\definecolor{darkgreen}{rgb}{0.1,0.7,0.1}
\definecolor{cyan}{rgb}{0.7,0.0,0.7}
\definecolor{dblue}{rgb}{0.2,0.2,0.8}
\definecolor{maroon}{rgb}{0.76,.13,.28}
\definecolor{burntorange}{rgb}{0.81,.33,0}
\definecolor{tealblue}{rgb}{0.212,0.459, 0.533}
\definecolor{myyellow}{rgb}{0.8627451 , 0.55294118, 0.2784314}

\definecolor{almostblack}{HTML}{111111}
\definecolor{mypink}{rgb}{0.93359375, 0.62109375, 0.83984375}

\definecolor{pp}{rgb}{0.43921569, 0.18823529, 0.62745098}
\definecolor{rr}{rgb}{0.5254902 , 0.00784314, 0.12941176}
\definecolor{bb}{rgb}{0.09019608, 0.23529412, 0.37647059}
\definecolor{yy}{rgb}{0.49803922, 0.3372549 , 0.0}
\definecolor{gg}{rgb}{0.02352941, 0.3372549 , 0.17647059}

\usepackage{amsmath,amsfonts,bm}

\def\1{\bm{1}}

\def\sp{space}

\def\rvr{{\mathbf{r}}}

\def\rmF{{\mathbf{F}}}

\def\vw{{\bm{w}}}

\def\vz{{\bm{z}}}

\def\mI{{\bm{I}}}

\def\mR{{\bm{R}}}
\def\mS{{\bm{S}}}
\def\mT{{\bm{T}}}

\DeclareMathAlphabet{\mathsfit}{\encodingdefault}{\sfdefault}{m}{sl}
\SetMathAlphabet{\mathsfit}{bold}{\encodingdefault}{\sfdefault}{bx}{n}
\newcommand{\tens}[1]{\bm{\mathsfit{#1}}}

\def\tZ{{\tens{Z}}}

\def\gD{{\mathcal{D}}}

\def\gL{{\mathcal{L}}}

\def\gT{{\mathcal{T}}}

\def\sR{{\mathbb{R}}}

\def\sZ{{\mathbb{Z}}}

\newcommand{\etens}[1]{\mathsfit{#1}}

\def\etB{{\etens{B}}}

\def\etD{{\etens{D}}}
\def\etE{{\etens{E}}}

\def\etU{{\etens{U}}}

\newcommand{\E}{\mathbb{E}}

\usepackage{symbols}
\usepackage{amsmath,amssymb}
\usepackage{enumitem}
\usepackage{graphicx}
\usepackage{booktabs}
\usepackage{tabularx}
\usepackage{multirow}       %
\usepackage{algorithm}
\usepackage{algorithmic}
\usepackage{bbm}
\usepackage{subcaption}
\usepackage{wrapfig}

\usepackage[most]{tcolorbox}

\usepackage{bbding}
\usepackage{pifont}

\definecolor{iccvblue}{rgb}{0.21,0.49,0.74}
\definecolor{lightcarminepink}{rgb}{0.9, 0.4, 0.38}
\usepackage[pagebackref,breaklinks,colorlinks,allcolors=iccvblue]{hyperref}

\usepackage[framemethod=tikz]{mdframed}
\mdfdefinestyle{MyFrame}{%
    linecolor=black,
    outerlinewidth=0.1pt,
    roundcorner=2pt,
    innertopmargin=0.3\baselineskip,
    innerbottommargin=0.3\baselineskip,
    innermargin = 0.2cm,
    outermargin = 0.cm,
    innerrightmargin=5pt,
    innerleftmargin=5pt,
    backgroundcolor=mybb!3!white}

\definecolor{mybrown}{rgb}{0.87058824, 0.56078431, 0.01960784}
\definecolor{myblue}{rgb}{0.3372549 , 0.70588235, 0.91372549}
\definecolor{mypurple}{rgb}{0.8, 0.47058824, 0.7372549 }
\definecolor{myorange}{rgb}{0.835, 0.368, 0}
\definecolor{mygreen}{rgb}{0.00784314, 0.61960784, 0.45098039}
\definecolor{mygt}{rgb}{0.0078125 , 0.57421875, 0.40625}
\definecolor{mysp}{rgb}{0.84765625, 0.515625  , 0.0234375}
\definecolor{mycitecolor}{rgb}{0,0.08,0.45}
\definecolor{mygr}{rgb}{0.9607,0.9607,0.9607}
\definecolor{myoo}{rgb}{0.992,0.9176,0.9019}
\definecolor{myrr}{HTML}{AE031A}
\definecolor{mybb}{HTML}{0155B3}

\usepackage{amsthm}
\newtheorem{claim}{Claim}
\newtheorem{hypothesis}{Hypothesis}

\def\paperID{} %
\def\confName{ICCV}
\def\confYear{2025}

\title{CLIPSym: Delving into Symmetry Detection with CLIP}

\author{
    Tinghan Yang \quad 
    Md Ashiqur Rahman \quad
    Raymond A. Yeh
    \\
    Department of Computer Science, Purdue University
}

\begin{document}

\maketitle

\begin{abstract}
Symmetry is one of the most fundamental geometric cues in computer vision, and detecting it has been an ongoing challenge. With the recent advances in vision-language models,~i.e., CLIP, we investigate whether a pre-trained CLIP model can aid symmetry detection by leveraging the additional symmetry cues found in the natural image descriptions. We propose CLIPSym, which leverages CLIP's image and language encoders and a rotation-equivariant decoder based on a hybrid of Transformer and $G$-Convolution to detect rotation and reflection symmetries. To fully utilize CLIP's language encoder, we have developed a novel prompting technique called Semantic-Aware Prompt Grouping (SAPG), which aggregates a diverse set of frequent object-based prompts to better integrate the semantic cues for symmetry detection. Empirically, we show that CLIPSym outperforms the current state-of-the-art on three standard symmetry detection datasets (DENDI, SDRW, and LDRS). Finally, we conduct detailed ablations verifying the benefits of CLIP's pre-training, the proposed equivariant decoder, and the SAPG technique. The code is available at \small{\url{https://github.com/timyoung2333/CLIPSym}}.
\end{abstract}

\section{Introduction}
Symmetry plays an important role in human perception and understanding of the world~\cite {driver1992preserved,delius1978symmetry,von1992dolphin,rodriguez2004symmetry,liu2013symmetry,funk20172017}. %
In computer vision, symmetry is one of the most fundamental geometric cues, providing essential information for tasks such as object recognition~\citep{vetter1994importance,pashler1990coordinate}, scene understanding~\citep{ecins2016cluttered,wilder2019local}, image matching~\citep{hauagge2012image}, and editing~\citep{lukavc2017nautilus}. Detecting symmetry, however, has been a long-standing question in computer vision due to the variations and complexities present in real-world scenarios~\cite {liu2010computational,tyler2003human,rauschert2011symmetry,je2024robust,seo2025leveraging,wu2024r2det}.

Earlier works relied on keypoint matching techniques~\citep{atadjanov2016reflection,shen2001robust,prasad2005detecting,sinha2012detecting,cicconet2017convolutional}, which involved comparing local descriptors of keypoints and their transformed counterparts. Although effective to some extent, these methods struggled with complex symmetry patterns or in the presence of noise.
More recently, deep learning-based approaches~\citep{funk2017beyond,seo2021learning,seo2022reflection} were proposed to detect reflection and rotation symmetries and have shown promising results. %
PMCNet~\citep{seo2021learning} proposed a method that relies on specially designed convolutional techniques rather than principled equivariant architectures, limiting its ability to consistently detect symmetry patterns across different orientations. Although EquiSym~\citep{seo2022reflection} addresses this limitation by leveraging group-equivariant convolutional networks, due to the limited availability of large-scale annotated symmetry datasets, the full potential of the learning based approach remains underexplored.

On the other hand, recent advances in pre-trained vision-language foundation models~\cite{radford2021learning,ramesh2021zero,jia2021scaling,yuan2021florence}, 
have shown remarkable generalization %
capabilities by leveraging large-scale datasets and joint training on visual and textual information. Image captions often contain words or phrases that carry the symmetry information of the object in the image. For example, in the case of internet-scale LAION-400M dataset~\citep{schuhmann2021laion}, around $10\%$ of the image captions contain words that convey shape/symmetry-related cues such as `rectangle,' `circle,' `oval', ~\etc. (see Appendix~\ref{sec:sup_laion} for detailed statistics). This observation suggests that vision-language models trained on such extensive image-text pairs are likely to contain useful symmetry cues in their learned text/image representations. A natural question arises: 
\vspace{-0.15cm}
\begin{quote}
\textit{How to leverage pre-trained vision-language models for symmetry detection?} 
\vspace{-0.15cm}
\end{quote}

In this paper, we present CLIPSym, a novel framework that leverages the pre-trained CLIP model for %
the detection of reflection and rotation symmetries in images. Our approach is motivated by the hypothesis that being trained on a large-scale vision-language dataset, %
the visual representations learned by CLIP contain %
knowledge that can benefit symmetry detection. 
The proposed CLIPSym contains CLIP's image and text encoders and introduces a decoder with guarantees on rotation equivariance. Given an input image, CLIPSym outputs a symmetry heatmap, where each pixel represents the probability of being a reflection axis or the rotation center. To fully leverage the potential of CLIP's language encoder, we propose a novel Semantic-Aware Prompt Grouping (SAPG) method, which aggregates multiple text prompts to enhance the model's understanding of symmetries. %

To validate our method's performance, we conduct a set of comprehensive experiments on three symmetry detection datasets and observe that our proposed CLIPSym achieves state-of-the-art performance for both reflection and rotation symmetry detection tasks. Finally, we analyze our model's equivariance properties and conduct ablations for each of the proposed components.

{\bf\noindent Our contributions are as follows:}
\begin{itemize}[topsep=0pt, leftmargin=16pt]
    \item We introduce CLIPSym, a framework that, for the first time, leverages the multimodal understanding abilities of CLIP 
    to achieve end-to-end detection of reflection and rotation symmetries.
    \item We propose SAPG, a novel prompting technique to enhance the model's understanding of symmetries through the aggregation of a diverse set of prompts.
    \item We propose a symmetry decoder with theoretical guarantees for rotation equivariance, which improves the model's robustness to diverse symmetry patterns.
    \item We demonstrate that CLIPSym achieves state-of-the-art performance across multiple benchmark datasets. Extensive ablation studies further validate the importance of CLIP pre-training, the SAPG technique, and the equivariant decoder.
\end{itemize}

\section{Related work}

{\bf\noindent Symmetry detection.} 
Earlier works tackled the task of symmetry detection primarily through \emph{keypoint matching}, which works by comparing local features of corresponding key points between an image and its mirrored version. Often, techniques such as spatial and angular auto-correlation are employed \citep{lin1997extracting,lee2009skewed,keller2006signal}. Local feature descriptors, such as SIFT \citep{seo2022reflection}, couture, and edge features \citep{atadjanov2016reflection,prasad2005detecting, wang2014unified,shen2001robust,wang2015reflection}, are frequently utilized to achieve a degree of equivariance to image transformations and detection of boundaries of symmetric objects.

Symmetry detection can also be formulated as a dense prediction task by assigning a score to each pixel of the image. \citet{tsogkas2012learning} employed a bag of features and multiple-instance learning in their model. On the other hand, \citet{gnutti2021combining} computed a symmetry score for each pixel using patch-wise correlation and gradient for validating candidate axes. \citet{funk2017beyond, fukushima2006symmetry} used data-driven learning-based approaches for symmetry detection. Polar matching convolution (PMC) \cite{seo2021learning} is used to attain higher reflection consistency in symmetry detection. To achieve perfect rotation and translation equivariance, \citet{seo2022reflection} used group equivariant CNNs to predict per-pixel symmetry scores.

While existing methods have achieved promising performance in symmetry detection, they still face challenges in modeling diverse symmetry patterns and lack large annotated datasets. In this paper, we aim to overcome these limitations by leveraging the power of the pre-trained CLIP model, which has learned visual-semantic representations with generalization capability to real-world scenes.

\vspace{2pt}
{\bf\noindent Equivariant networks.}
Equivariance to geometric transformations in input images constitutes a vital inductive bias, fostering improved generalization and consistency, particularly under conditions of limited training data. While Convolutional Neural Networks (CNNs) inherently exhibit equivariance to the translation operations, achieving equivariance to a broader spectrum of geometric transformations is not guaranteed. This broader family of equivariance is achievable through Group Equivariant CNNs \citep{cohen2016group} and parameter sharing strategies~\citep{kawano2021group, yeh2022equivariance,yeh2019chirality}. Notably, Steerable CNNs \citep{cohen2016steerable, weiler2018learning, weiler20183d, weiler2019general} offer an efficient approach by representing filters in terms of steerable bases. Recent works have extended the scope of equivariance to include diverse transformations, such as scaling~\citep{worrall2019deep, rahman2024truly, sosnovik2019scale}, sampling \cite{rahman2025group, rojas2022learnable}, color changes~\citep{lengyel2024color}, permutation~\cite{ravanbakhsh_sets,zaheer2017deep,hartford2018deep,yeh2019chirality,liu2021semantic,
liu2020pic,yeh2019diverse}, and extending beyond CNN architectures to encompass Vision Transformers~\citep{rojas2023making, xu20232}. Equivariance is particularly important in our task of symmetry detection, as it allows the model to consistently identify symmetrical patterns regardless of their orientation %
or position in the image, leading to more robust and accurate predictions. %

\vspace{2pt}
{\bf\noindent Vision \& language models.} CLIP~\citep{radford2021learning}, a seminal pre-trained vision-language model, has gathered significant attention and has been widely adopted in various downstream tasks, including monocular depth estimation~\citep{Hu_2024_WACV}, sound source localization~\citep{Park_2024_WACV}, scene text detection and spotting~\citep{xue2022language}, video understanding~\citep{rasheed2023fine}, semantic segmentation~\citep{luddecke2022image}, etc. Recently, prompting has emerged as a prominent paradigm for efficiently adapting pre-trained models to downstream tasks. \citet{zhou2022learning} and \citet{zhou2022conditional} propose methods to automatically learn prompt tokens that yield strong performance on target tasks. \citet{khattak2023maple} introduce a multi-modal prompting approach to effectively adapt CLIP to various applications. Furthermore, \citet{bahng2022exploring} explores the use of visual prompts to probe CLIP's visual representation learning capabilities.

\begin{figure*}[t]
    \centering
    \includegraphics[width=0.94\linewidth]{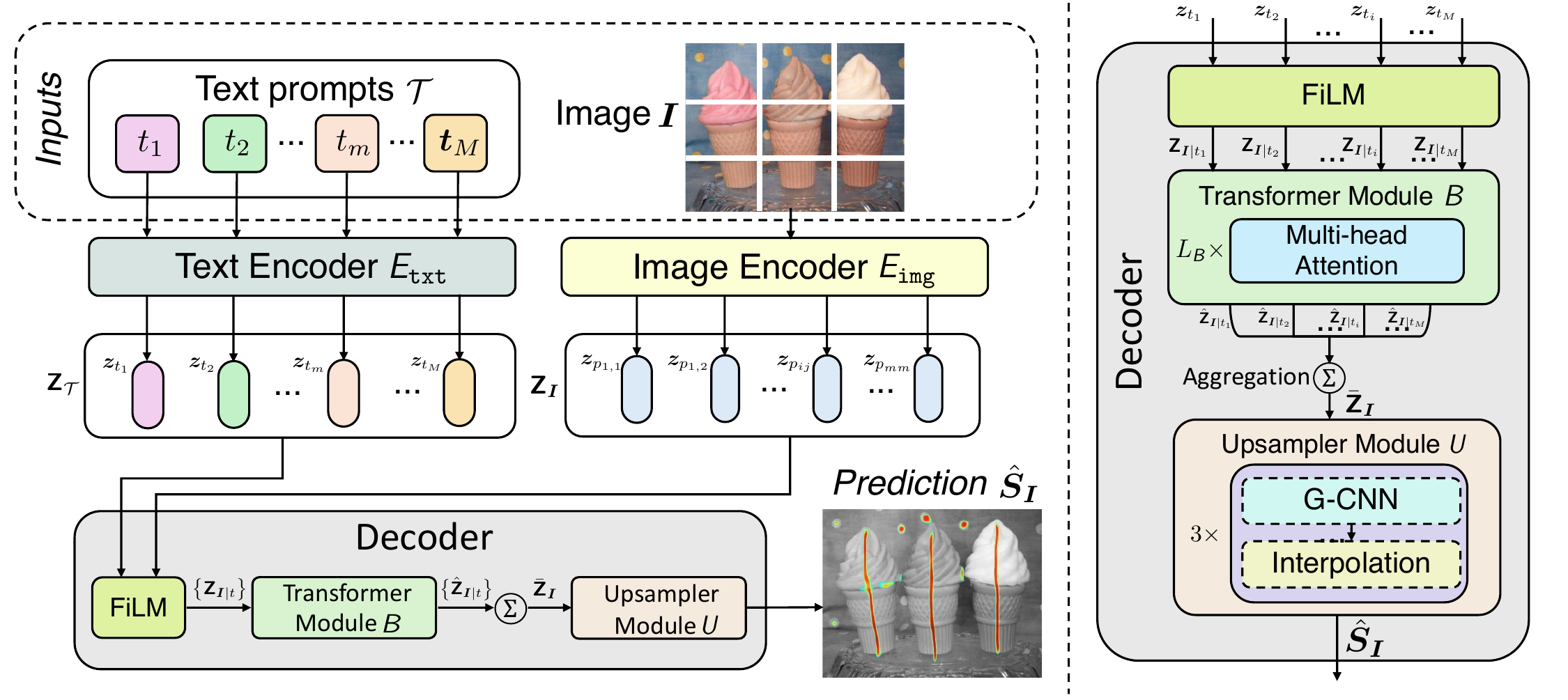}
    \vspace{-0.2cm}
    \caption{Overview of the proposed CLIPSym architecture. {\textbf{Left:}} The text encoder $\etE_{\tt txt}$ encodes prompts in set $\gT$ as $\tZ_{\gT}$ and the image encoder $\etE_{\tt img}$ encodes patches in image $\mI$ as $\tZ_\mI$. $\tZ_\mI$ and $\tZ_\gT$ are then mixed and aggregated in the decoder to get the final predicted symmetry heatmap $\hat{S}_{\mI}$. {\textbf{Right}: Visualization of decoder details.}
    }
    \vspace{-0.2cm}
    \label{fig:pipeline}
\end{figure*}

\section{Approach}\label{sec:app}

We propose CLIPSym, a model that leverages the pre-trained CLIP model (ViT-B/16) for the task of symmetry detection. Given an input of an image $\mI \in \sR^{ H \times W \times 3}$, CLIPSym outputs the predicted symmetry heatmap $\hat{\mS}_\mI \in [0,1]^{H \times W}$, which represents the probability of each pixel being reflection axes or the rotation center for objects in %
$\mI$.
As CLIP has been trained on an internet-scale dataset, we hypothesize that such pre-training would be beneficial to symmetry detection. The main challenge is how to build a model that utilizes this pre-trained knowledge. 

To leverage the image information from CLIP, we use the pre-trained image encoder $\etE_{\tt img}$ to extract image features from the set of image patches tokens $\tZ_{\mI} =  \{\vz_{p_{ij}}\}$, where $\vz_{p_{ij}} \in \sR^d$ denotes the image feature of patch $p_{ij}$ at position $(i,j) \in \sZ_M^2$, with $M$ representing the number of patches along each dimension. 
To leverage the text information from CLIP, we design SAPG, which integrates a set of text prompts $\gT = \{t_1, t_2, \dots \}$ and use the text encoder $\etE_{\tt txt}$ to extract a set of text tokens $\tZ_{\gT} = \{\vz_{t_1}, \vz_{t_2}, \dots \}$ where each $\vz_{t_i} \in \sR^d$. 

With the image and text tokens extracted, we then propose a \textit{rotation equivariant decoder} to mix and aggregate the tokens into a final heatmap. A visual overview of our approach is illustrated in~\figref{fig:pipeline}. 
We will now discuss the decoder details in~\secref{subsec:decoder}, followed by the prompting technique SAPG in~\secref{subsec:prompt}, and training details in~\secref{subsec:training}.

\subsection{Rotation equivariant decoder}%
\label{subsec:decoder}
{\bf\noindent Decoder architecture.}
The decoder module $\etD$ takes the set of the image tokens $\tZ_\mI$ and text tokens $\tZ_\gT$ as inputs and generates the final symmetry heatmap \ie,  $\hat{\mS}_\mI$; an overview is provided in~\figref{fig:pipeline} (left). Our proposed decoder module consists of three modules, namely, a FiLM block, a Transformer module followed by aggregation, and finally, a rotation equivariant upsampler. We design the decoder to be rotation equivariant, as prior work~\cite{seo2022reflection} has shown equivariance guarantees to benefit the performance of symmetric detection. We now discuss each of the building blocks.

\textit{}\textit{\ding{172} FiLM block:} 
A FiLM~\citep{dumoulin2018feature} conditioning layer utilizes the text tokens to modulate the image features, allowing image tokens to carry textual semantic information. For each text token $\vz_{t} \in \tZ_\gT$, the FiLM layer generates a set of image tokens modulated by text condition $t$:
\bea
&\tZ_{\mI|t} &= {\tt FiLM}(\vz_t, \tZ_\mI) \\
&&= \{\vz_{p_{ij}|t}|(i, j) \in \sZ_M^2 ~,~ \vz_{p_{ij}|t} \in \sR^d\},
\eea
where each $\vz_{p_{ij}|t}$ is computed as
\bea \label{eqn:flim} \vz_{p_{ij}|t} = \gamma(\vz_t) \odot \vz_{p_{ij}} + \beta(\vz_t). \eea
Here, $\odot$ denotes element-wise multiplication between text and image patch features, and $\gamma(\cdot), \beta(\cdot)$ are linear layers.

\textit{\ding{173} Transformer module \& aggregation:} With the set of image tokens modulated for each text $t$, we then use a Transformer module $\etB$ to further learn the spatial dependencies between patches, which is crucial for detecting global symmetry structures. The Transformer module consists of several multi-headed attention blocks, each containing a self-attention layer and multi-layer perceptron (MLP) layers followed by layer normalization as described in ViT~\citep{dosovitskiy2020image}.

Each set of text-modulated image tokens $\tZ_{\mI|t}$ is passed to the transformer module $\etB$ to obtain the set of updated tokens
\bea
\hat{\tZ}_{\mI|t} = \{\hat{\vz}_{p_{ij}|t} | (i, j) \in \sZ_M^2 \} = \etB(\tZ_{\mI|t})~\forall t \in \gT.
\eea
Next, we aggregate across all text prompts to construct the set of final tokens $\bar\tZ_{\mI}$ via a weighted average: 
\bea\nonumber
\bar\tZ_{\mI} = \{\bar{\vz}_{p_{ij}} | (i, j) \in \sZ_M^2 \}\\ ~~\text{where each } \bar{\vz}_{p_{ij}} = \sum_{t \in \gT} \vw_t \hat{\vz}_{p_{ij}|t}.
\eea
Here, $\vw_t \in \sR$ is a weight scalar corresponding to prompt $t$ therefore $\vw \in \sR^{|\gT|}$ learns to combine the patch-conditioned tokens. The weights satisfy $\vw_t \ge 0$ and $\sum_{t=1}^{|\gT|} \vw_t = 1$. The upsampler will next process these tokens.

\textit{\ding{174} Rotation equivariant upsampler:} To achieve equivariance, we choose to use steerable $G$-Conv~\citep{cohen2016steerable} and choose $G$ to be roto-translation group $\sZ_M^2 \rtimes C_n$, where $C_n$ denotes a group of $360^\circ/n$ rotations and $n = 4k, \text{where } k \in \mathbb{Z}^+$.

As $G$-Conv takes a feature map on a group as input, we first need to convert the set of aggregated tokens $\bar\tZ_{\mI}$ to a grid and then lift it to the roto-translation group. Recall, each element $\bar\vz_{p_{ij}} \in \bar\tZ_{\mI}$ has a corresponding spatial location $p_{ij}$. We put back ({$\tt Grid$}) the elements into a 2D feature map as:
\bea\nonumber
\rmF \triangleq {\tt Grid} (\bar\tZ_{\mI}) \in \sR^{d \times M \times M}\\ \text{~~where~~} \rmF[:,i,j] = \bar\vz_{p_{ij}}~~\forall~(i,j) \in \sZ_M^2.
\eea
We then lift this feature map $\rmF$ to the roto-translation group. The lifted feature map $\rmF^{\uparrow} \in \sR^{|C_n| \times d' \times m \times m }$ is defined as
\bea
\rmF^{\uparrow} \triangleq {\tt Concat}\left([\mR_\theta \rmF; ~\forall \theta \in C_n]\right),
\eea
where $\mR_\theta$ denotes rotation on $2D$ plane. In more details,
\bea
  \rmF^\uparrow[i, \theta, x, y] = \rmF[i,x',y']\\  \text{ where } 
\begin{pmatrix}
x' \\
y'
\end{pmatrix}
=
\begin{pmatrix}
\cos\theta & \sin\theta \\
-\sin\theta & \cos\theta
\end{pmatrix}
\begin{pmatrix}
x \\
y
\end{pmatrix}.
\eea
More compactly, we denote this action as $[x',y'] = \rvr_{-\theta} (x,y)$.

The lifted feature map $\rmF^{\uparrow}$ is then passed through $3$ layers of $G$-Conv~\cite{cohen2016group} and 
 $4\times$ %
 bi-linear upsampling,%
where $G$-conv computes the following:
\bea\nonumber
(\rmF^\uparrow \star_{G} \psi)[\theta, x, y] = \sum_{\theta' \in G} \sum_{(x',y') \in \sZ_M^2} \\
\rmF^\uparrow[\theta', x', y'] \psi[\theta'- \theta, \rvr_{-\theta}[(x'-x, y'-y) ].
\eea 
Finally, the feature on the roto-translation group is mean-pooled along the rotation dimension $\theta$ in the last layer to generate the final prediction symmetry heatmap $\hat{\mS}_\mI$.

\vspace{2pt}
{\bf\noindent Rotation equivariance guarantees.} We will now show that our proposed decoder $\etD$ is rotation equivariant to the group $C_4$, \ie, a 2D ``rotation'' on the image tokens $\tZ_\mI$ leads to the same rotation of the prediction $\hat\mS_\mI$. We define a ``rotation'' using the action $\mT_\theta$ on the set of patch-features $\tZ_\mI$ as 
\bea\label{eq:transform}
 \mT_\theta \tZ_\mI \triangleq \{ \mT_\theta \vz_{p_{ij}} \} = \{ \vz_{p_{\pi_\theta(ij)}}\} ,
\eea 
where $\pi_\theta$ rotates the $2D$ coordinates,~\ie, a permutation on the patch location $(i,j)$.

\begin{mdframed}[style=MyFrame,align=center]
\begin{claim}\vspace{-0.2cm}
The decoder $\etD$ is rotation ($\mT_\theta, \mR_\theta$)-equivariant to $C_4$, \ie,
\vspace{-0.2cm}
\bea
\etD( \mT_\theta \tZ_\mI, \tZ_\gT) = \mR_\theta \hat{\mS}_\mI~~\forall\theta \in C_4.
\eea
\end{claim}
\end{mdframed}
\vspace{-0.5cm}
\begin{proof}
It is sufficient to prove that each component of the decoder is equivariant. 

\textit{\ding{172} FiLM block:} The FiLM block performs element-wise affine transformation (multiplication and addition) for each of the patch features separately. As any operation performed individually on each element of the set is permutation equivariant~\cite{zaheer2017deep}, \,  i.e., ${\tt FiLM}(\mT_\theta \tZ_\mI, \vz_t) = \mT_\theta \tZ_{\mI|t}$.

\textit{\ding{173} Transformer module \& aggregation:}
In $C_4$, $\theta \in \{ n \cdot 90^\circ: n \in \sZ \}$ then $\pi_\theta$ acts as a permutation on the patch location $(i,j)$ and the action $\mT_\theta$ can be described as a permutation on the patch features.
\bea{\tt Grid}( \mT_\theta \tZ_\mI) = \mR_\theta  [{\tt Grid}(\tZ_\mI)].
\eea

Transformer layers are equivariant to permutation on the order of the tokens~\cite{yun2019transformers}. So, the application of transformer block $\etB$ is equivariant, \ie $\etB( \mT_\theta \tZ_{\mI|t}) = \mT_\theta  \etB(\tZ_{\mI|t})$. 
The aggregation of tokens $\bar\tZ_{\mI}$ is constructed by a weighted average over the text prompts. As the tokens are spatially aligned, the aggregated token remains equivariant.

\textit{\ding{174} Rotation equivariant upsampler:}
The upsampler $\etU$ and relative interpolations are equivariant to $C_n$, where $n$ is a multiple of 4 by design. As $C_4$ is a subgroup of $C_n$, the upsampler is equivariant to $C_4$, \ie,
\bea\etU({\tt reshape}(\mT_\theta\bar\tZ_{\mI}))=\etU(\mR_\theta \rmF)=\mR_\theta \etU(\rmF) = \mR_\theta \hat{\mS}_\mI. \eea
This concludes the proof.
\end{proof}

\subsection{Semantic-Aware Prompt Grouping (SAPG)}\label{subsec:prompt}

While commonly used text prompts such as ``{\tt a photo of a [CLASS]}'' seems to be a good choice, symmetry is a highly abstract concept that almost exists across a variety of objects, making it unlikely for CLIP's pre-training data to include specific descriptions like ``symmetry axes'' or ``rotation centers''. 
In CLIPSym, we propose a novel prompting technique SAPG to address this challenge. SAPG constructs a set of prompts $\gT = \{t_1, t_2, \cdots, t_M\}$, with each prompt $t_m$ representing a string of the combination of $K$ frequent object classes that appear in the dataset, which are separated by spaces. Formally: %
\bea
    t_m = ``[{\tt obj}_{m_1}]~~[{\tt obj}_{m_2}]~~\cdots~~[{\tt obj}_{m_K}]",
\eea
where ${\tt obj}_{m_k}$ represents the $k$-th object class in the $m$-th prompt. For example, with $K=3$ objects in each prompt, the prompt $t_m$ can be ``{\tt apple cloud table}''.  Note that we use the same prompt set \(\gT\) for all images, ensuring a consistent semantic initialization across the dataset.
More details of prompts are provided in Appendix~\ref{sec:sup_prompt}. %

The design of SAPG is motivated by three key insights:
\begin{itemize}
    \item \emph{Better initialization via frequent objects:}  
Since pre-trained CLIP has good language-image alignment, using frequent objects as prompts leads the model to focus more on regions where symmetry is naturally present, thus allowing the model to learn the underlying symmetric structures rather than dealing with noisy or inconsistent semantic signals from less common words.
    
\item \emph{Aggregation for prompts:}
Grouping multiple prompts allows the model to leverage complementary semantic cues such that the model can capture broader aspects of symmetry than a single prompt, which typically focuses on only limited aspects of symmetry. Moreover, the aggregated embeddings of grouped prompts are refined during training, which provides a more robust representation of symmetry.

\item \emph{Fixed prompts for a universal concept:}  
As the concept of symmetry is universal, this means its core characteristics do not vary significantly from one image to another, it is reasonable to use a fixed set of prompts rather than adapting them for each image. In this way, the model has a more consistent semantic anchor that reflects symmetry. Moreover, although prompts are fixed, their embeddings are continuously updated during training, allowing them to gradually evolve to capture the essential characteristics of symmetry more accurately.
\end{itemize}

In our experiments, we explore various strategies for selecting object classes. Detailed prompt design and more discussions on the motivations of language are in Appendix~\ref{sec:sup_prompt} and~\ref{sec:sup_discussion}.

\subsection{Model training}\label{subsec:training}
In symmetry detection, the class imbalance problem arises due to the low ratio of foreground pixels indicating the rotation/reflection axis to the background pixel. To address this issue, we follow prior works~\citep{lin2017focal, seo2021learning} to utilize the $\alpha-$focal loss defined as
\bea\label{eq:focal_loss}
    \gL_{\text{focal}}(\mI) = \sum\limits_{x, y} -{\alpha'_{\mI_{xy}}}(1 - \hat{\mS}'_{\mI_{xy}})^\lambda \log(\hat{\mS}'_{\mI_{xy}}),
\eea
where $\hat{\mS'}_{\mI_{xy}}$ represents the predicted heatmap for image $\mI$ at position $(x, y)$, $\alpha'_{\mI_{xy}}$ represents the symmetry/non-symmetry class balance factor calculated from a pre-defined scalar $\alpha$, and $\lambda$ denotes the focusing parameter. Detailed definitions can be found in the Appendix~\ref{sec:sup_imp}.

We fine-tune from the pre-trained CLIP text and image encoders. %
This decision is driven by two key considerations. First, CLIP has been pre-trained on a vast corpus of image-text pairs, but its training objective does not specifically focus on symmetry detection. %
Second, the prompts are aimed at capturing the abstract concept of symmetry rather than specific object classes. Hence, it requires fine-tuning the text encoder to map these prompts to text tokens for symmetry detection.

\section{Experiments}
 For a fair comparison, we strictly follow the evaluation protocol of prior works~\cite{seo2022reflection,seo2021learning}. We first discuss the experimental setup, followed by the results, and conclude with a set of ablation studies. 
 
\subsection{Experimental setup}

{\bf\noindent Dataset.}
As in prior works~\cite{seo2022reflection}, we conduct experiments on the task of reflection and rotation symmetry detection using three datasets: DENDI~\cite{seo2022reflection}, SDRW~\cite{seo2021learning}, and LDRS~\cite{seo2021learning}.

The DENDI dataset consists of 2493 and 2079 images annotated for reflection axes and rotation centers, with 1750/374/369 and 1459/313/307 images in train/validation/test splits for reflection and rotation symmetry, respectively. On the other hand, SDRW and LDRS are reflection datasets that have 51/-/70 and 1110/127/240 images in train/validation/test splits. Although the original SDRW dataset includes both rotation and reflection data, we only use its reflection data because its rotation data has already been incorporated into DENDI.

\vspace{2pt}
{\bf\noindent Baselines.} We compare our approach with three baseline methods considered by~\citet{seo2022reflection}, including SymResNet~\citep{funk2017beyond}, PMCNet~\citep{seo2021learning}, and EquiSym~\citep{seo2022reflection}. SymResNet applied ResNet~\citep{he2016deep} to detect reflection and rotation symmetries using human-labeled annotated data. PMCNet proposed polar matching convolution to detect reflection symmetries by leveraging polar feature pooling and self-similarity encoding. EquiSym introduced a group-equivariant convolutional network to detect reflection and rotation by utilizing equivariant feature maps, surpassing the performance of all previous methods. 

Beyond existing works, we further consider additional baselines: 
CLIPSym$^{\text{no-text}}$ only uses a CLIP image encoder followed by the same \emph{equivariant decoder} as CLIPSym without any text conditioning, which will reflect the benefits of language.
CLIPSym$^{\text{scratch}}$ trains the proposed model from scratch instead of using the pre-trained CLIP, which aims to show that pre-training is helpful. CLIPSym$^{\text{non-eq.}}$ is a variant of CLIPSym with a non-equivariant decoder that uses standard CNN blocks for upsampling, which will show the importance of the design of the equivariant decoder.

\vspace{2pt}
{\bf\noindent Evaluation metrics.}
To evaluate the symmetric detection tasks, we report the F1-score following~\citet{seo2022reflection}. We also report robustness and consistency metrics with respect to rotation and reflection transformations for each of the models. The evaluation metrics are summarized below: 
    
\emph{F1-score} ($\uparrow$) is formally calculated as 
\bea
    \textnormal{F1} = \max\limits_{\tau}\Big(\frac{2 \cdot \text{precision}_{\tau} \cdot \text{recall}_{\tau}}{\text{precision}_{\tau} + \text{recall}_{\tau}}\Big),
\eea
where $\tau \in [0,1]$ is used to threshold the predicted score map at various levels in the range of $[0, 1]$ to obtain binary maps. For each threshold, we compute the F1-score by comparing the predictions against the ground-truth binary heatmaps at the pixel level. The max F1-score across all thresholds is reported as the final performance measure.
     
     {\it Robustness-score} assesses the model's robustness under transformations, including reflections and rotations, which is calculated as the F1-score on the transformed dataset. During the assessment of rotation robustness, we sample rotation angles uniformly distributed between $[-45^{\circ}, 45^{\circ}]$ and apply them to images in the dataset and their relative ground-truth heatmaps. For reflection robustness, we randomly apply a horizontal flip on each image.

    {\it Consistency-score} is defined as the cross-entropy loss between the transformed model's outputs and the model's output on the transformed input images. Formally, %
    \bea\label{eq:cons}
    \textnormal{Consistency} = \frac{1}{\lvert\gD\rvert}\sum_{\mI \in \gD} {\E}_{\mT} \bigg[ \text{CE}\big(\mT(\hat{\mS}_{\mI}), \hat{\mS}_{\mT(\mI)} \big)\bigg],
    \eea
    where $\mT$, defined in~\equref{eq:transform}, denotes the transformation (\eg, rotation or reflection), CE denotes the cross-entropy function between the two symmetry heatmaps. %
    A lower score indicates a higher consistency, suggesting the model can maintain more consistent predictions faced with reflection or rotation transformations. 

\vspace{2pt}
{\bf\noindent Implementation details.}
As the backbone network, we adopt the pre-trained CLIP model~\citep{radford2021learning} with a ViT-B/16 structure. The model is trained for 500 epochs using the Adam optimizer. To meet the input requirements of the image encoder, training images are reshaped to $417\times417$ resolution by resizing the original images while maintaining the aspect ratio and padding if necessary. During testing, images are reshaped using the same process, where the predictions are cropped and resized to the original image sizes before computing metrics. See Appendix~\ref{sec:sup_imp} for more details.

\subsection{Results}

\setlength{\tabcolsep}{3pt}
\begin{table}[t]
    \centering
    \small
    \begin{tabular}{l ccc}
    \specialrule{.15em}{.05em}{.05em}
        {\bf{Method}} & \bf Pre-training &\bf{Reflection F1} & \bf{Rotation F1} \\
        \hline
        SymResNet$^*$~\citep{funk2017beyond} & ImageNet & 30.7 & 11.9 \\
        PMCNet$^*$~\citep{seo2021learning} & ImageNet & $52.0$ & -- \\
        PMCNet~\citep{seo2021learning} & ImageNet & $53.8 \pm 0.5$ & -- \\
        EquiSym$^*$~\citep{seo2022reflection} & ImageNet & \underline{$64.5$} & \underline{22.5} \\
        EquiSym~\citep{seo2022reflection} & ImageNet & $61.7 \pm 0.6$ & {$22.0 \pm 0.7$} \\
        \midrule
        $\text{CLIPSym}^{\text{no-text}}$ & ImageNet & $54.8 \pm 0.2$ & $9.0 \pm 0.1$ \\
        $\text{CLIPSym}^{\text{no-text}}$ & CLIP & $63.7 \pm 0.3$ & $17.7 \pm 0.2$ \\
        $\text{CLIPSym}^{\text{scratch}}$& -- & $32.1 \pm 0.2$ & $4.7 \pm 0.2$ \\
        $\text{CLIPSym}^{\text{non-eq.}}$& CLIP & $62.9 \pm 0.2$ & \underline{$24.2 \pm 0.1$} \\
        $\text{CLIPSym}^{\text{eq.}}$& CLIP & {$\bf{66.5 \pm 0.2}$} & ${\bf 25.1 \pm 0.1}$ \\
    \specialrule{.15em}{.05em}{.05em}
    \end{tabular}
        \caption{Quantitative comparison of F1-score (\%) on the DENDI dataset~\cite{seo2022reflection}. Results of SymResNet$^*$, PMCNet$^*$, and EquiSym$^*$ are obtained from the EquiSym~\citep{seo2022reflection} paper, while PMCNet and EquiSym are reproduced using the publicly available code. As SymResNet~\citep{funk2017beyond} does not have publicly available code, we are unable to report its standard deviation.}
        \vspace{-0.2cm}
    \label{tab:dendi}
\end{table}

{\bf\noindent Quantitative results.}
In~\tabref{tab:dendi} and~\tabref{tab:pmc_test}, we present the F1-score of baseline models pre-trained on different datasets or trained from scratch for detecting reflection and rotation symmetries on DENDI and reflection symmetry on SDRW and LDRS datasets, respectively. 

\noindent From~\tabref{tab:dendi} and~\tabref{tab:pmc_test}, we observe the following:

 {\it CLIPSym achieves SOTA performance.} In~\tabref{tab:dendi}, we observe that CLIPSym has the highest F1 score across both tasks, outperforming EquiSym$^*$ by 2.0\% and 2.6\%, the previous SOTA, on the DENDI dataset.
   
    {\it CLIP's pre-training is helpful.} In both~\tabref{tab:dendi} and~\tabref{tab:pmc_test}, we observe that CLIPSym pre-trained on CLIP significantly outperforms CLIPSym trained from scratch. CLIPSym without text conditioning pretrained on CLIP also outperforms those pretrained on ImageNet, which suggests that pre-training on a larger and more diverse dataset is beneficial.

     {\it CLIPSym effectively leverages the information from the text encoder.} This can be seen from the comparison between CLIPSym$^\text{no-text}$ and CLIPSym, where CLIPSym outperforms its counterpart in all settings. This suggests that the text encoder provides additional contextual information that helps the model to understand symmetries better.

\begin{table}[t]
\small
\setlength{\tabcolsep}{1pt}
    \centering
    \begin{tabular}{lcccc}
    \specialrule{.15em}{.05em}{.05em}
        {\bf{Method}} & \bf{Pre-training} & \bf{SDRW F1} & \bf{LDRS F1} & \bf{Mixed F1} \\
        \hline
        PMCNet~\citep{seo2021learning} & ImageNet & $40.8 \pm 0.4$ & $30.5 \pm 0.5$ & $33.8 \pm 0.2$ \\
        EquiSym~\citep{seo2022reflection} & ImageNet & \underline{$48.2 \pm 0.1$} & \underline{$37.7 \pm 0.1$} & \underline{$41.1 \pm 0.1$} \\
        \midrule
        CLIPSym$^\text{no-text}$ & ImageNet & ${31.3 \pm 0.1}$ & ${25.3 \pm 0.1}$ & ${27.0 \pm 0.1}$ \\
        CLIPSym$^\text{no-text}$ & CLIP & ${46.8 \pm 0.2}$ & ${36.2 \pm 0.1}$ & ${39.7 \pm 0.1}$ \\
        CLIPSym$^\text{scratch}$ & -- & ${10.8 \pm 0.3}$  & ${10.4 \pm 0.2}$ & ${10.8 \pm 0.3}$ \\
        CLIPSym$^\text{non-eq.}$ & CLIP & {${47.8 \pm 0.3}$} & {${37.0 \pm 0.1}$} & ${40.8 \pm 0.2}$ \\
        CLIPSym$^\text{eq.}$ & CLIP & {$\bf{51.8 \pm 0.3}$} & {$\bf{39.5 \pm 0.1}$} & {$\bf{42.8 \pm 0.1}$} \\
    \specialrule{.15em}{.05em}{.05em}
    \end{tabular}
        \vspace{-0.2cm}
        \caption{F1-score of reflection symmetry detection on SDRW, LDRS, and their mixed datasets. %
    }
            \vspace{-0.2cm}
    \label{tab:pmc_test}
\end{table}

\begin{table}[t]
    \setlength{\tabcolsep}{3pt}
    \centering
    \small
    \begin{tabular}{lccc}
    \specialrule{.15em}{.05em}{.05em}
    {\bf{Method}} & {\bf{Pre-training}} & \bf{Robustness ${\bm \uparrow}$} & \bf{Consistency ${\bm \downarrow}$} \\ %
    \hline
    PMCNet~\citep{seo2021learning} & ImageNet & 52.2 & 0.417 \\
    EquiSym~\citep{seo2022reflection} & ImageNet & 57.1 & 0.244 \\ %
    CLIPSym$^\text{non-eq.}$ & CLIP & \underline{58.3} & \underline{0.093} \\
    CLIPSym$^\text{eq.}$ & CLIP & {\bf 59.7} & {\bf0.082} \\
    \specialrule{.15em}{.05em}{.05em}
    \end{tabular}
    \vspace{-0.15cm}
    \caption{Equivariance robustness and consistency evaluation results for DENDI reflection dataset under $[-45^\circ, 45^\circ]$ uniformly distributed rotation operations. 
    }
    \vspace{-0.5cm}
    \label{tab:con_rob}
    \end{table}

Beyond performance, we further study how equivariance plays a role in the models. In~\tabref{tab:con_rob}, we present the Consistency and Robustness score of models on the DENDI reflection dataset. Here, we report the consistency and robustness of rotations uniformly randomly sampled within $\pm45$ degrees. Interestingly, we observe CLIPSym$^\text{non-eq.}$ with a non-equivariant decoder surpasses the two compared baselines in consistency and robustness. Note that both EquiSym~\citep{seo2022reflection} and our CLIPSym use the $C_8$ group-equivariant convolutions, which are only equivariant at intervals of $45^{\circ}$. 
This again highlights the importance of CLIP's pre-training in the encoder for consistent image representations. Finally, the full CLIPSym with an equivariant decoder further improves the consistency and robustness of the model. Please refer to Appendix~\ref{sub_sec:sup_cons_robust} for consistency and robustness results on the SDRW and LDRS datasets.

\begin{table}[t]
    \setlength{\tabcolsep}{3pt}
    \centering
    \small
\begin{tabular}{lccc}
    \specialrule{.15em}{.05em}{.05em}
{\bf{Method}} & {PMCNet~\citep{seo2021learning}} & {EquiSym~\citep{seo2022reflection}} & {CLIPSym (ours)} \\ %
\hline
{\bf{GFLOPs}} & 167.7 & {\bf 114.0} & 148.8 \\
    \specialrule{.15em}{.05em}{.05em}
\end{tabular}
\vspace{-0.15cm}
\caption{Comparison of computation cost in GFLOPs.} 
\vspace{-0.5cm}
\label{tab:comp_cost}
\end{table}

\vspace{2pt}
{\bf \noindent Comparisons on the computational cost.}
We report the computational costs in GFLOPs as in~\tabref{tab:comp_cost} for each of the baselines and our method. We observe that CLIPSym has a slightly higher computational cost at 148.8 GFLOPs compared to EquiSym (114.0 GFLOPs), but is more efficient than PMCNet (167.7 FLOPs). That is, CLIPSym achieves a significant performance improvement over other baselines at a moderate increase in computation.
\begin{figure*}[htb]
  \vspace{-0.1cm}
  \centering
  \begin{subfigure}[t]{0.49\textwidth}
  \begin{minipage}[t]{.244\linewidth}
  \vspace{0pt}
    \centering
    \caption*{\scalebox{1.0}{Ground Truth}}
      {\includegraphics[width=\linewidth,trim={0 1.07cm 0 0cm},clip]{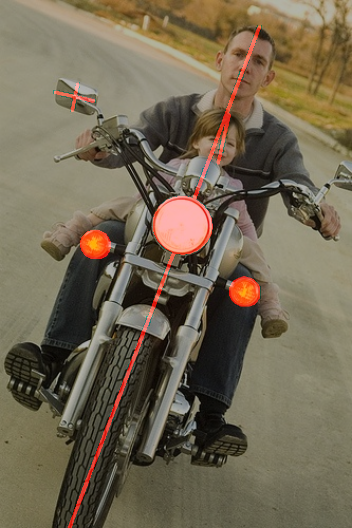}}
      {\includegraphics[width=\linewidth]{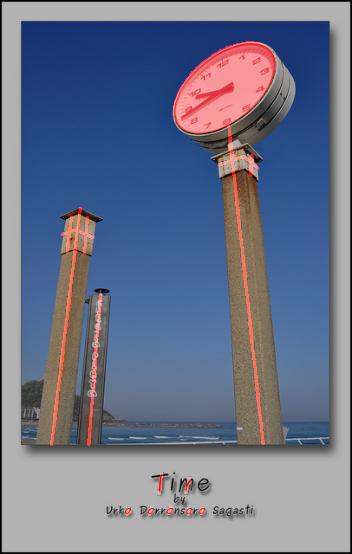}}
      {\includegraphics[width=\linewidth,trim={0 2cm 0 0},clip]{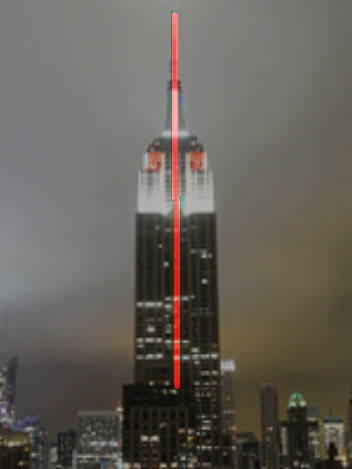}}
  \end{minipage}
  \begin{minipage}[t]{.244\linewidth}
  \vspace{0pt}
    \centering
    \caption*{\scalebox{1.0}{PMCNet~\citep{seo2021learning}}}
      {\includegraphics[width=\linewidth,trim={0 1.07cm 0 0cm},clip]{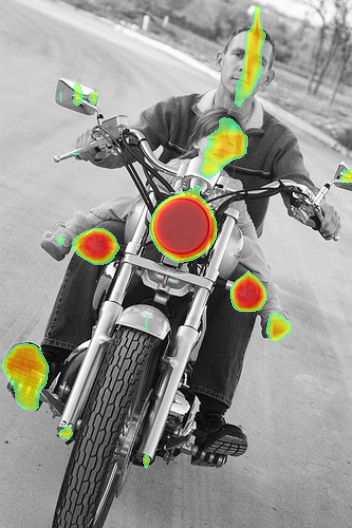}}
      {\includegraphics[width=\linewidth]{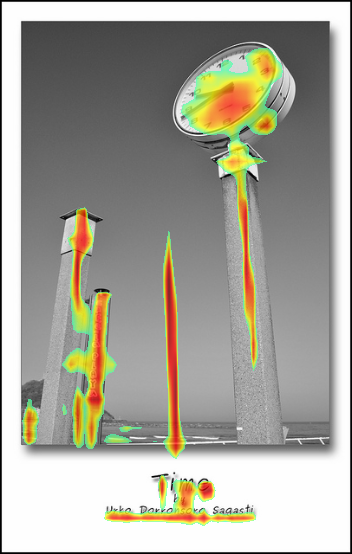}}
      {\includegraphics[width=\linewidth,trim={0 1.83cm 0 0},clip]{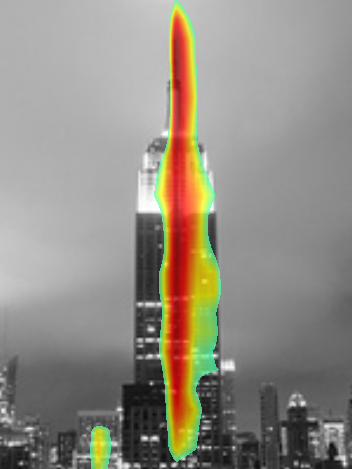}}
    \label{fig:vis_ref_pmc}
  \end{minipage}
    \begin{minipage}[t]{.244\linewidth}
    \vspace{0pt}
    \centering
    \caption*{\scalebox{1.0}{EquiSym~\citep{seo2022reflection}}}
      {\includegraphics[width=\linewidth,trim={0 1.07cm 0 0cm},clip]{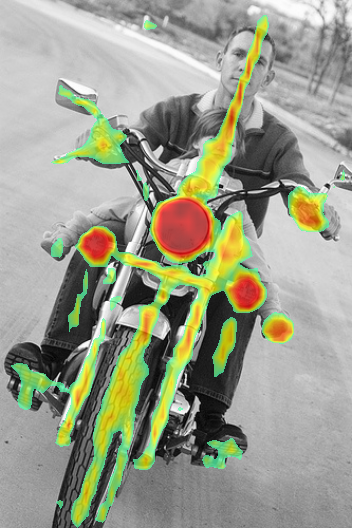}}
      {\includegraphics[width=\linewidth]{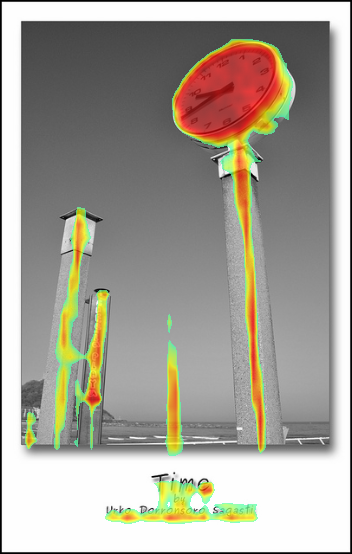}}
      {\includegraphics[width=\linewidth,trim={0 1.83cm 0 0},clip]{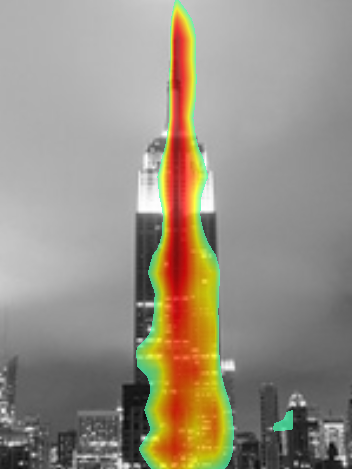}}
  \end{minipage}
      \begin{minipage}[t]{.244\linewidth}
      \vspace{0pt}
    \centering
    \caption*{\scalebox{1.0}{CLIPSym}}
      {\includegraphics[width=\linewidth,,trim={0 1.07cm 0 0cm},clip]{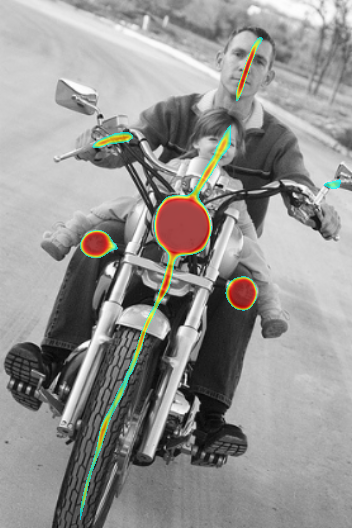}}
      {\includegraphics[width=\linewidth]{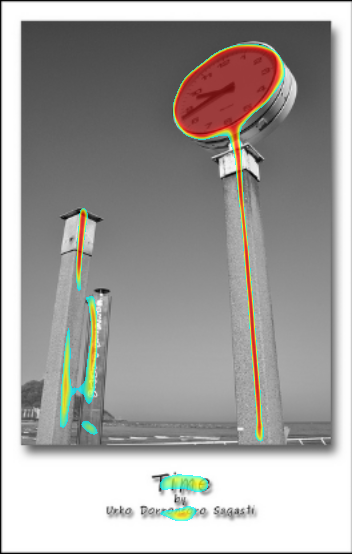}}
      {\includegraphics[width=\linewidth,trim={0 1.83cm 0 0},clip]{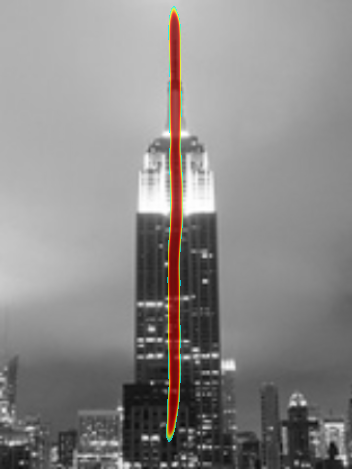}}
      \label{fig:vis_ref_clipsym}
  \end{minipage}
  \vspace{-0.45cm}
\caption{Reflection detection results on DENDI-\textit{ref}.}
\label{fig:vis_ref}
\end{subfigure}
\hspace{-0.05mm}
\begin{subfigure}[t]{0.485\textwidth}
  \centering
  \begin{minipage}[t]{.3259\linewidth}
  \vspace{0pt}
    \centering
    \caption*{\scalebox{1.0}{Ground Truth}}
      {\includegraphics[width=\linewidth]{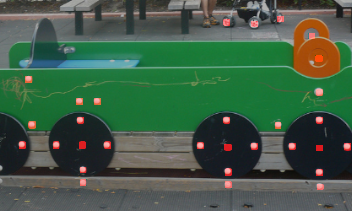}}
      {\includegraphics[width=\linewidth]{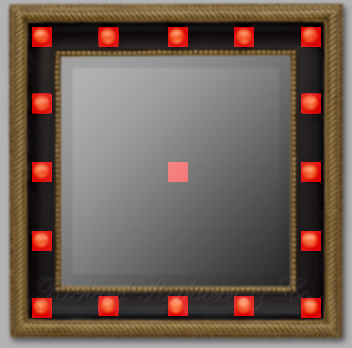}}
      {\includegraphics[width=\linewidth]{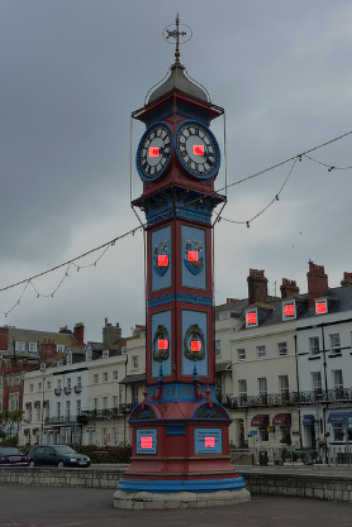}}
    
  \end{minipage}
  \begin{minipage}[t]{.3259\linewidth}
  \vspace{0pt}
    \centering
    \caption*{\scalebox{1.0}{EquiSym~\citep{seo2022reflection}}}
      {\includegraphics[width=\linewidth]{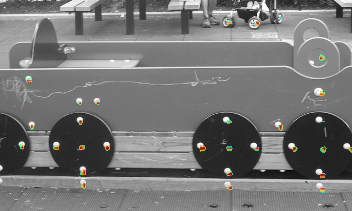}}
      {\includegraphics[width=\linewidth]{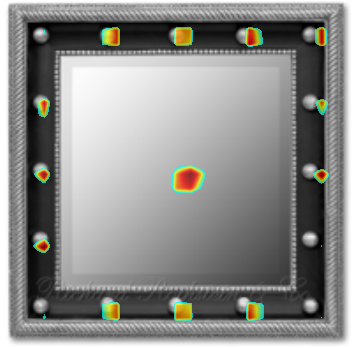}}
      {\includegraphics[width=\linewidth]{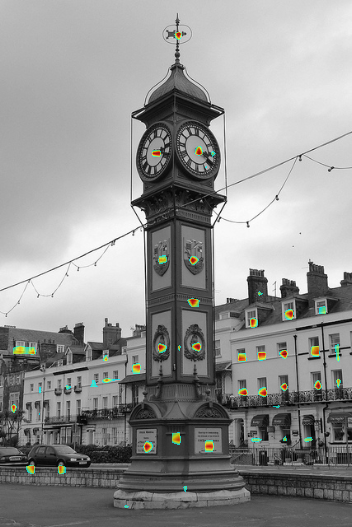}}
    \label{fig:rot_eq}
  \end{minipage}
    \begin{minipage}[t]{.3259\linewidth}
    \vspace{0pt}
    \centering
    \caption*{\scalebox{1.0}{CLIPSym}}
      {\includegraphics[width=\linewidth]{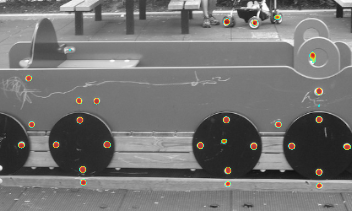}}
      {\includegraphics[width=\linewidth]{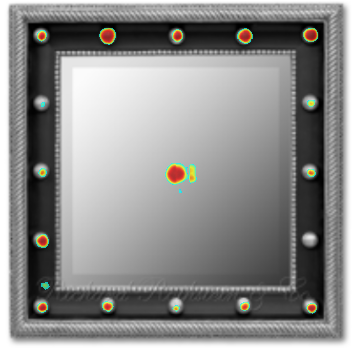}}
      {\includegraphics[width=\linewidth]{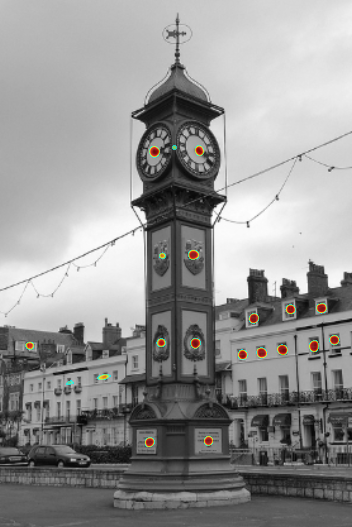}}
  \end{minipage}
  \vspace{-0.45cm}
  \caption{Rotation detection results on DENDI-\textit{rot}.}
  \label{fig:vis_rot}
  \end{subfigure}
        \vspace{-0.1cm}
\caption{Visualization of the reflection and rotation symmetry detection on the DENDI dataset.
}
\end{figure*}

\vspace{2pt}
{\bf\noindent Qualitative results.}
In~\figref{fig:vis_ref} and~\figref{fig:vis_rot}, we compare the predicted reflection and rotation heatmaps of different models and the ground truth. %
We observe that CLIPSym generates sharper and more accurate symmetry heatmaps compared to the baselines.

\begin{figure*}[t]
    \vspace{-0.1cm}
    \centering
    \begin{subfigure}[t]{1.0\textwidth}
    \begin{minipage}[t]{.137\linewidth}
        \vspace{0pt}
        \centering
        \caption*{\scalebox{1.}{Original image}}
        {\includegraphics[width=\linewidth]{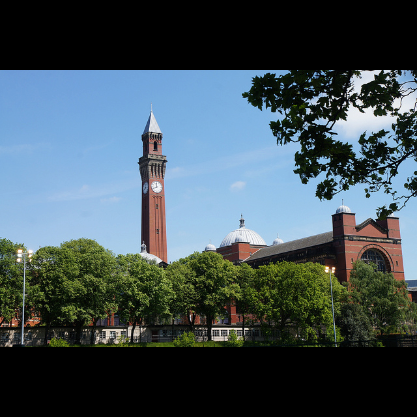}}
        {\includegraphics[width=\linewidth]{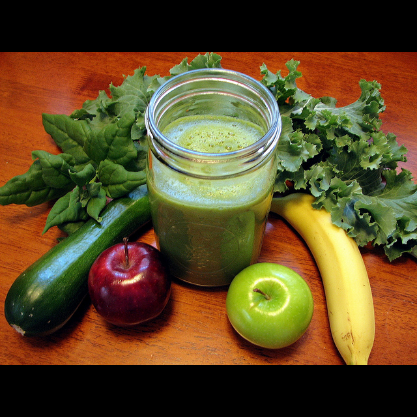}}
        {\includegraphics[width=\linewidth]{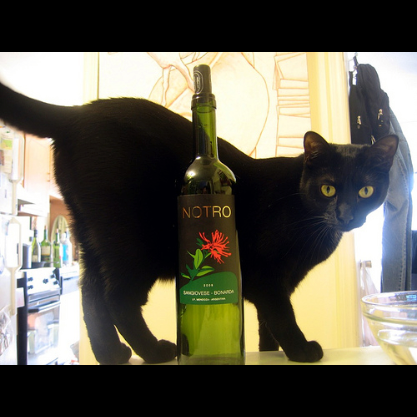}}
    \end{minipage}
    \begin{minipage}[t]{.137\linewidth}
        \vspace{0pt}
        \centering
        \caption*{\scalebox{1.}{Ground Truth (GT)}}
        {\includegraphics[width=\linewidth]{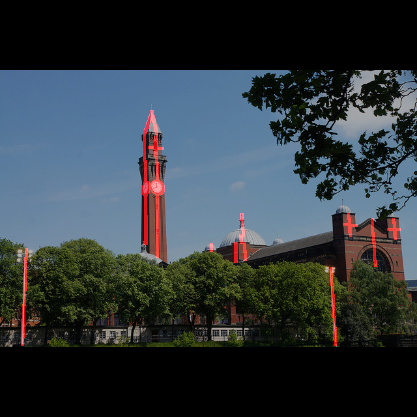}}
        {\includegraphics[width=\linewidth]{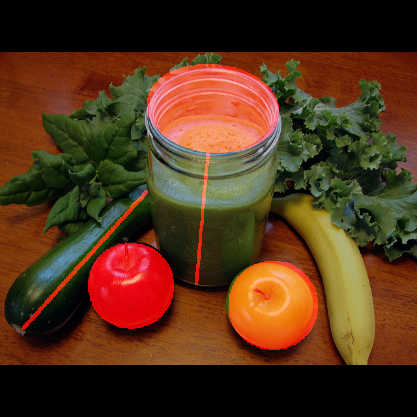}}
        {\includegraphics[width=\linewidth]{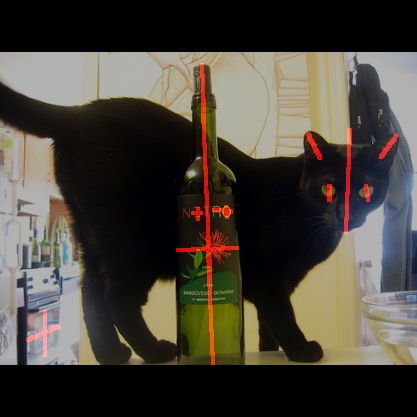}}
    \end{minipage}
    \begin{minipage}[t]{.137\linewidth}
        \vspace{0pt}
        \centering
        \caption*{\scalebox{1.}{Rotated GT}}
        {\includegraphics[width=\linewidth]{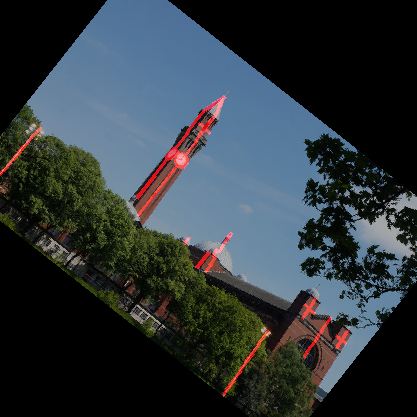}}
        {\includegraphics[width=\linewidth]{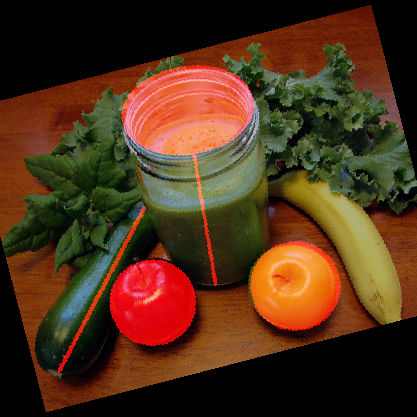}}
        {\includegraphics[width=\linewidth]{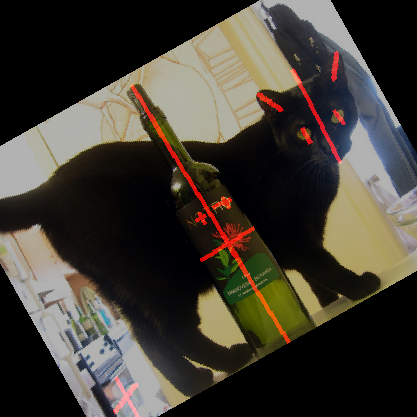}}
    \end{minipage}
    \begin{minipage}[t]{.137\linewidth}
        \vspace{0pt}
        \centering
        \caption*{\scalebox{1.}{EquiSym $\hat{\mS}_{\mT(\mI)}$}}
        {\includegraphics[width=\linewidth]{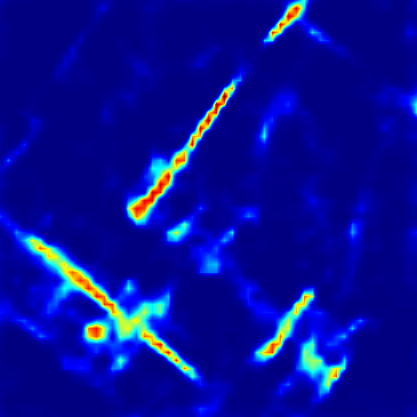}}
        {\includegraphics[width=\linewidth]{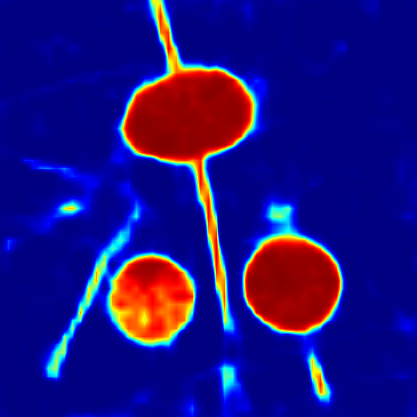}}
        {\includegraphics[width=\linewidth]{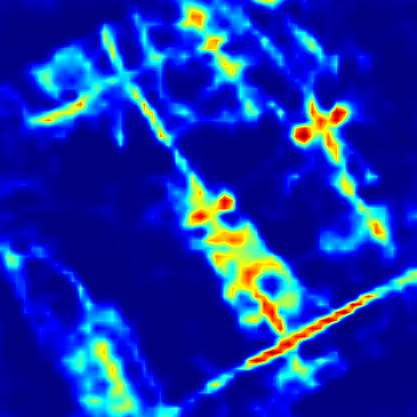}}
    \end{minipage}
    \begin{minipage}[t]{.137\linewidth}
        \vspace{0pt}
        \centering
        \caption*{\scalebox{1.}{EquiSym $\mT(\hat{\mS}_\mI)$}}
        {\includegraphics[width=\linewidth]{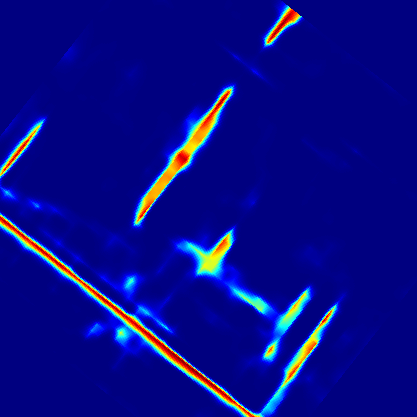}}
        {\includegraphics[width=\linewidth]{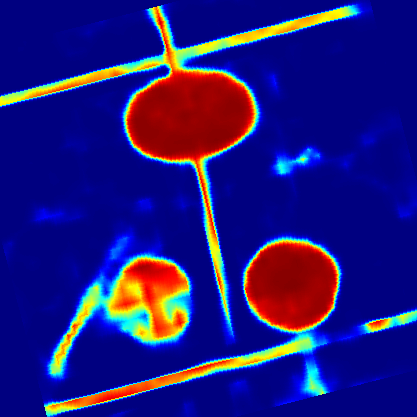}}
        {\includegraphics[width=\linewidth]{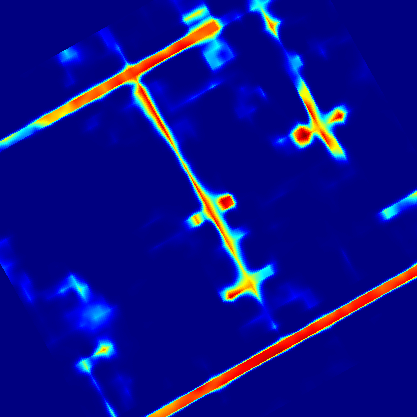}}
    \end{minipage}
    \begin{minipage}[t]{.137\linewidth}
        \vspace{0pt}
        \centering
        \caption*{\scalebox{1.}{CLIPSym $\hat{\mS}_{\mT(\mI)}$}}
        {\includegraphics[width=\linewidth]{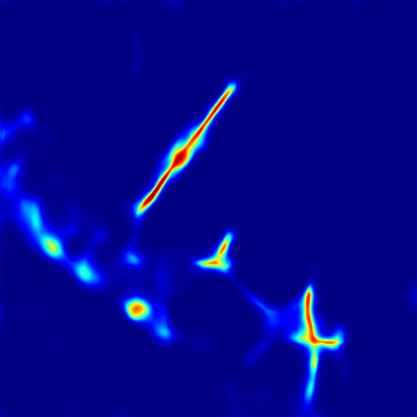}}
        {\includegraphics[width=\linewidth]{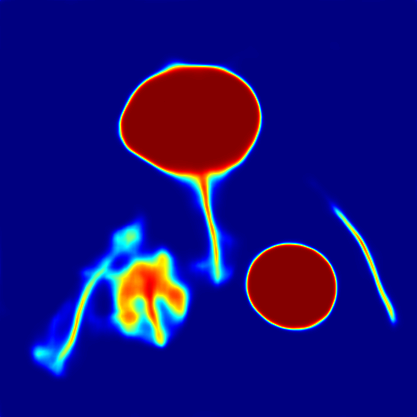}}
        {\includegraphics[width=\linewidth]{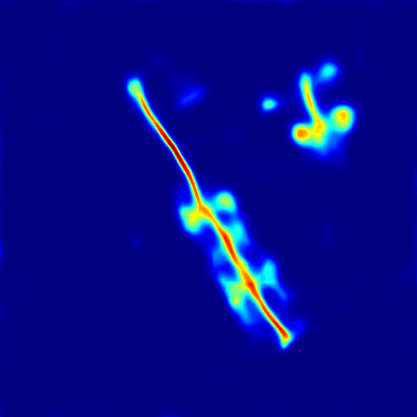}}
    \end{minipage}
    \begin{minipage}[t]{.137\linewidth}
        \vspace{0pt}
        \centering
        \caption*{\scalebox{1.}{CLIPSym $\mT(\hat{\mS}_\mI)$}}
        {\includegraphics[width=\linewidth]{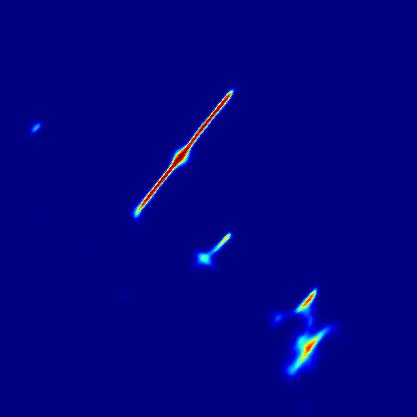}}
        {\includegraphics[width=\linewidth]{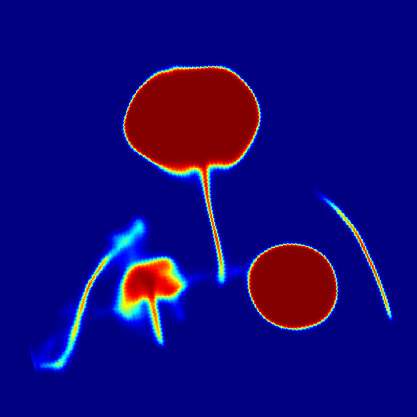}}
        {\includegraphics[width=\linewidth]{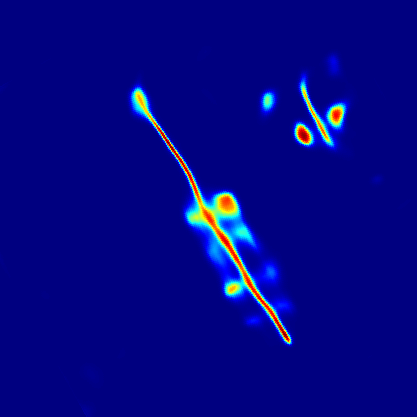}}
    \end{minipage}
  
  \end{subfigure}
  \caption{Examples of the original image, ground truth, rotated ground truth, EquiSym and CLIPSym's predicted heatmaps $\hat{\mS}_{\mT(\mI)}$ on the rotated image and the rotated heatmap $\mT(\hat{\mS}_\mI)$. Observe that CLIPSym's results are more consistent under rotation transformations.
  }
  \label{fig:cons_vis_rot}
  \end{figure*}

In~\figref{fig:cons_vis_rot}, we present heatmaps of EquiSym and CLIPSym, which take images under random rotation transformations within $[-45^\circ, 45^\circ]$ as inputs to illustrate the model's robustness and consistency. We observe that even though EquiSym is a fully equivariant model, %
CLIPSym generates more consistent heatmaps. This is because end-to-end equivariant models using steerable filters require exact symmetry at the input. They are not guaranteed to be equivariant when there are interpolation artifacts, cropping of the image, or when the rotation is not a multiple of $90^\circ$. On the contrary, CLIP's image encoder is robust to such transformations due to large-scale training and generates consistent image features. Our equivariant decoder module $\etD$, generates a consistent symmetry heatmap from CLIP's image feature.

\figref{fig:cons_vis_rot} also shows that compared with EquiSym, CLIPSym's predictions are sharper and contain less noise, suggesting that CLIPSym is more robust. 
More results are shown in~\figref{fig:sup_vis_rot}.

\subsection{Ablation studies}
{\bf\noindent Prompt initialization.}
In~\tabref{tab:multi_prompts}, we investigated the impact of different prompt initialization methods for the CLIPSym text encoder on the reflection symmetry detection performance using the DENDI dataset. We explored two main categories of prompt initialization: single prompt ($M=1$) and multiple prompts ($M>1$).

\begin{table}[t]
\setlength{\tabcolsep}{6pt}
\centering

\resizebox{\linewidth}{!}{%
\begin{tabular}{llc}
    \specialrule{.15em}{.05em}{.05em}
& \bf{Prompt context} & \bf{Ref. F1} \\
\hline
\multirow{4}{*}{\bf{Single prompt}}
& \textit{A single phrase containing K tokens} & \\
& ``reflection axis'' & $64.4$ \\
& ``symmetry axes in the image'' & 64.8 \\
& frequent object classes ($K$=25) & $\bf{65.8}$ \\
\midrule
\multirow{6}{*}{\bf{Multi-prompt}} & \textit{$M$ prompts, each with $K$ tokens} & \\
& $M=25, K=1$ & 65.3 \\
& $M = 25, K = 4$ & ${\bf 66.5}$ \\
& $M = 25, K = 16$ & 65.9 \\
& $M = 50, K = 4$ & 65.4 \\
& $M = 50, K = 16$ & 64.4 \\
    \specialrule{.15em}{.05em}{.05em}
\end{tabular}
}
\vspace{-0.1cm}
\caption{Ablation results of different prompt initialization methods for CLIPSym text encoder on DENDI reflection dataset. As defined in~\secref{subsec:prompt}, $M$ represents the number of prompts, and $K$ represents how many words there are in each prompt.}
\vspace{-0.1cm}

\label{tab:multi_prompts}

\end{table}

For a single prompt, we evaluated using arbitrary phrases or sentences, such as ``reflection axis'' and ``symmetry axis'', which achieved F1 scores of 64.4 and 64.8, respectively. We also tested combinations of words with frequent objects (65.3). Furthermore, using multiple prompts containing $M$ tokens each consistently outperforms single prompt methods.%

We find that using 25 prompts with 4 tokens each yields the highest F1 score of 66.5, demonstrating the effectiveness of leveraging multiple diverse prompts for initialization. 
These results highlight the importance of careful prompt engineering and show that utilizing multiple semantically relevant prompts can improve performance. Appendix~\ref{sec:sup_prompt} provides more detailed descriptions of the prompts. 

\vspace{2pt}
{\bf\noindent Trainable components.} In~\tabref{tab:trainable_clip}, we investigate the impact of making the text encoder and image encoder trainable in the proposed CLIPSym. The best performance is achieved when both encoders are trainable, which suggests that both encoders contribute to the symmetry detection task. When the image encoder is frozen, whether the text encoder is trainable or not, the performance drops significantly. This suggests that the image encoder plays a more crucial role in symmetry detection than the text encoder. 

\vspace{2pt}
{\bf \noindent Different CLIP models. } Beyond using the ViT-B/16 model as in the main experiments, we also experimented with other variants of CLIP models, including ViT-L/14,  SigLIP~\citep{zhai2023sigmoid} which replaces the softmax loss with a sigmoid loss for improved feature separability, and MetaCLIP~\citep{xu2023demystifying} which created a balanced and noise-reduced dataset to improve the training of CLIP model. 
As reported in~\tabref{tab:clips}, we observe that MetaCLIP achieves even better reflection detection performance than our model (66.7 vs. 66.5), and SigLIP achieves slightly worse performance (65.8). This suggests that CLIPSym has the potential to be further improved as more advanced CLIP backbones are developed.

\begin{table}
\hspace{-0.3cm}
\begin{minipage}{.47\linewidth}
\centering
    \centering
        \resizebox{\linewidth}{!}{%
    \begin{tabular}{ccc}
    \specialrule{.15em}{.05em}{.05em}
    \multicolumn{2}{c}{\bf{Trainable Encoder}} & \multirow{2}{*}{\bf{Ref. F1}} \\
    \textit{Text} & \textit{Image} & \\
    \hline
    \xmark & \xmark & 59.4 \\
    \cmark & \xmark & 58.9 \\
    \xmark & \cmark & \underline{65.3} \\
    \cmark & \cmark & {\bf{66.5}} \\
    \specialrule{.15em}{.05em}{.05em}
    \end{tabular}
    }
    \caption{F1-scores evaluated on DENDI reflection dataset under different settings. 
    }
    \vspace{-0.3cm}
    \label{tab:trainable_clip}
\end{minipage}
\hspace{6pt}
\begin{minipage}{.51\linewidth}

    \setlength{\tabcolsep}{3pt}
    \centering
    \small
    \resizebox{\linewidth}{!}{%
\begin{tabular}{lc}
    \specialrule{.15em}{.05em}{.05em}
{\bf{CLIP Version}} & {\bf{Ref. F1}} \\ %
\hline
CLIP/ViT-B-16 & {66.5 $\pm$ 0.2} \\
CLIP/ViT-L-14 & {65.4 $\pm$ 0.2} \\
SigLIP/ViT-B-16 & {65.8 $\pm$ 0.3} \\
MetaCLIP/ViT-B-16 & {\bf{66.7 $\pm$ 0.3}} \\
    \specialrule{.15em}{.05em}{.05em}
\end{tabular}
}
\caption{Comparison of different versions of CLIP model on reflection symmetry detection on DENDI.}
    \vspace{-0.3cm}
\label{tab:clips}

\end{minipage}
\end{table}

\section{Conclusion}
In this paper, we introduce CLIPSym, a new approach for symmetry detection that builds on the pre-trained CLIP model. Leveraging CLIP's powerful generalization and cross-modal capabilities, our method adapts it specifically for symmetry detection through prompt learning to capture geometrically relevant features. Additionally, the proposed equivariant decoder module boosts the model's robustness and consistency against random transformations. Our approach achieves state-of-the-art performance across all evaluated datasets.

\clearpage
\bibliographystyle{abbrvnat}
\bibliography{main}

\clearpage
\section*{Appendix}
\setcounter{section}{0}
\renewcommand{\theHsection}{A\arabic{section}}
\renewcommand{\thesection}{A\arabic{section}}
\renewcommand{\thetable}{A\arabic{table}}
\setcounter{table}{0}
\setcounter{figure}{0}
\renewcommand{\thetable}{A\arabic{table}}
\renewcommand\thefigure{A\arabic{figure}}
\renewcommand{\theHtable}{A.Tab.\arabic{table}}%
\renewcommand{\theHfigure}{A.Abb.\arabic{figure}}%
\renewcommand\theequation{A\arabic{equation}}
\renewcommand{\theHequation}{A.Abb.\arabic{equation}}%

\newcommand{\visspace}{\scalebox{1.75}[1]{\textvisiblespace}}

\begin{itemize}[topsep=0pt, leftmargin=16pt]
\item In~\secref{sec:sup_prompt}, we introduce more details about the structure of SAPG and provide several examples of the object classes and the set of prompts that are used in our experiments.
\item In~\secref{sec:sup_discussion}, we present a more in-depth explanation of this paper's motivation. Specifically, we compute statistics on the presence of symmetry cues within the vision-language dataset, analyze the benefits of language in symmetry detection from a theoretical perspective, and further discuss why SAPG works better.
\item In~\secref{sec:sup_res}, we present additional ablation studies and visualization results.
\item In~\secref{sec:sup_imp}, we provide additional implementation details.

\end{itemize}

\section{Structure of text prompts}\label{sec:sup_prompt}
To construct language prompts for symmetry detection, we use Grounded-SAM~\citep{ren2024grounded} to extract the frequent 2081 object classes from the DENDI dataset. The full list will be provided in the released code. Below we show the first 100 objects.
\begin{mdframed}[style=MyFrame,align=center]
\emph{man, pole, stand, white, building, sit, table, floor, sky, person, red, street sign, food, traffic sign, road, clock, plate, green, attach, catch, sign, park, peak, street corner, tree, platter, woman, car, stop sign, blue, tower, black, play, lush, blanket, yellow, road sign, stool, bell tower, grass, curb, tray, field, walk, stare, cloudy, pavement, ball, child, dinning table, photo, water, boy, ride, spire, animal, girl, drive, brown, fill, vegetable, cat, fly, footstall, room, hand, sea, lay, cup, container, pillar, flower, city, beverage, motorcycle, grassy, bowl, license plate, wear, fruit, shirt, countertop, dog, snow, plane, lamp, rail, motorbike, home appliance, toy, stone building, electronic, bus, chair, swinge, pizza, racket, tennis racket, rural, vase}
\end{mdframed}

Therefore, if we want to construct a set of prompts $\gT$ with $M=3$ prompts and each containing $K=3$ words, considering using the first $3\times 3=9$ objects as an example, $\gT$ can be construct as follows:
\begin{align*}
\gT = \{\underbrace{\textit{"man pole stand"}}_{t_1}, \underbrace{\textit{"white building sit"}}_{t_2}, \\
\underbrace{\textit{"table floor sky"}}_{t_3}\}
\end{align*}
Note, the ``frequent objects ($K=25$)'' row in~\tabref{tab:multi_prompts} uses the first 25 objects in the above list.

As stated in Sec.~\ref{subsec:prompt}, during training, the prompts are fixed and shared across all images. Our goal is not to search for the optimal prompt because the searching space is extremely large, but to show the grouping structure is helpful. We further discuss the benefits of this design in Sec.~\ref{sec:sup_sapg_better}.

\section{Discussions on the impacts of language}\label{sec:sup_discussion}
\subsection{Language cues about symmetry in CLIP's pre-training}\label{sec:sup_laion}
\begin{table}[]
\centering
\begin{tabular}{l l}
\toprule
\textbf{Shape/Symmetry}      &\textbf{Occurrence} \\ 
      \textbf{Word}              &   (\%)\\
\midrule
Ring & 4.2718 \\
Line & 1.9806 \\
Arc & 1.5185 \\
Ball & 1.4913 \\
Square & 0.4095 \\
Oval & 0.1699 \\
Cone & 0.1606 \\
Arrow & 0.1572 \\
Circle & 0.1518 \\
Globe & 0.0892 \\
Rectangle & 0.0830 \\
Cube & 0.0776 \\
Grid & 0.0708 \\
Pyramid & 0.0685 \\
Triangle & 0.0592 \\
Spiral & 0.0503 \\
Sphere & 0.0413 \\
Cylinder & 0.0324 \\
Hexagon & 0.0277 \\
Crescent & 0.0274 \\
Prism & 0.0163 \\
Octagon & 0.0109 \\
Checkerboard & 0.0070 \\
Helix & 0.0066 \\
Pentagon & 0.0050 \\
Ellipse & 0.0035 \\
Rhombus & 0.0025 \\
Trapezoid & 0.0021 \\
Torus & 0.0007 \\
Semicircle & 0.0006 \\
Dodecahedron & 0.0005 \\
Tetrahedron & 0.0005 \\
Icosahedron & 0.0004 \\
Parallelogram & 0.0004\\
\bottomrule
\end{tabular}
\caption{Percentage of image captions containing shape/symmetry related words in the LAION-400M dataset.}
\label{tab:laion_stat}
\end{table}

LAION-400M~\citep{schuhmann2021laion} is a large-scale public dataset containing $400M$ image-caption pairs, which is potentially similar to the dataset that CLIP was trained on. In~\tabref{tab:laion_stat}, we use GPT-4o to generate a few symmetry and shape-related words and calculate the percentage of the occurrence of these words within LAION-400M's captions. We observe that common shape-associated words such as `ring,' `line,' and `ball' have appeared more frequently than complex geometric shapes such as `parallelogram,' `icosahedron,' and `tetrahedron.' Nevertheless, as the dataset is very large, even the occurrence of $0.0004\%$ translates to the presence of $1600$ image-caption pair containing complex shape concepts such as `icosahedron.' Pre-training on such diverse image-caption pairs enables the CLIP model to learn image representations that encode rich symmetry-related information.

\subsection{A theoretical perspective on the benefits of language}\label{sec:sup_language_better}
In this subsection, we provide a theoretical perspective to analyze the benefits of using language in the symmetry detection task.

\begin{hypothesis}
\label{hyp:language}
Suppose there exists a perfect image encoder $\etE^*_{\tt img}$ which leads to the best visual features for image $\mI \in \sR^{ H \times W \times 3}$, i.e.,
$\tZ^*_\mI = \etE^*_{\tt img}(\mI)$.
Provided that language contains cues about symmetry, we assume the best visual features are offset by an additive term $\delta^*(t)$ that depends on a language prompt $t$, plus zero-mean noise $\varepsilon_{\mI}$:
\bea\vspace{-0.2cm}
\label{eq:fix}
\tZ_\mI = \tZ^*_\mI - \delta^*(t) + \varepsilon_{\mI},
\eea
where $\sE[\varepsilon_{\mI}] = 0$ and $\delta^*(t) \ne 0$ when language provides symmetry cues.
\end{hypothesis}
Then we make a claim that language is beneficial under Hypothesis~\ref{hyp:language}:
\begin{mdframed}[style=MyFrame,align=center]
\begin{claim}\vspace{-0.2cm}
Using language prompt $t$, a FiLM layer of the form (following equation~\ref{eqn:flim})
\bea
\label{eq:sup_film}
\tZ_{\mI|t} = \gamma(z_t) \odot \tZ_\mI + \beta(z_t),
\eea
with elementwise multiplication $\odot(\cdot)$ and trainable linear mappings $\gamma(\cdot), \beta(\cdot) \in \sR^d$, can reduce the expected error of visual features. Formally,
\bea
\sE\bigl[\|\tZ_{\mI|t} - \tZ^*_\mI\|\bigr] < \sE\bigl[\|\tZ_\mI - \tZ^*_{\mI}\|\bigr]
\eea
\end{claim}
\end{mdframed}

\begin{proof}
According to~\ref{eq:fix}, the ideal additive fix to $\tZ_\mI$ is 
\bea
f^*(\tZ_\mI) = \tZ_\mI + \delta^*(t),
\eea
where $f(\cdot)$ represents a function which modulates $\tZ_\mI$ to fit for the symmetry detection task, and $f^*(\cdot)$ correpsondingly represents the best fix function.

We then show FiLM can implement $f*(\cdot)$. Simply choose
\bea
\gamma(z_t) = \mathbf{1} (\text{all ones vector}), \;\; \beta(z_t) = \delta^*(t)
\eea
and apply to~\ref{eq:sup_film}, then
\bea
\tZ_{\mI|t} = \tZ_{\mI} + \delta^*(t) = \tZ^*_{\mI} + \varepsilon_{\mI}.
\eea
Hence, the language modulated visual features differ from the best visual features $\tZ^*_\mI$ by only $\varepsilon_{\mI}$ rather than $\delta^*(t) - \varepsilon_{\mI}$. Therefore,
\bea
\sE\bigl[\|\tZ_{\mI|t} - \tZ^*_\mI\|^2\bigr] & = \sE\bigl[\|\varepsilon_\mI\|^2\bigr],
\eea
while
\begin{align}
\sE\bigl[\|\tZ_\mI - \tZ^*_{\mI}\|^2\bigr] & = \sE\bigl[\|\delta^*(t) - \varepsilon_\mI\|^2\bigr] \\
& = \sE\bigl[\|\varepsilon_\mI\|^2\bigr] + \sE\bigl[\|\delta^*(t)\|^2\bigr] - 2\sE\bigl[\varepsilon_\mI \delta^*(t)\bigr] \\
& = \sE\bigl[\|\varepsilon_\mI\|^2\bigr] + \sE\bigl[\|\delta^*(t)\|^2\bigr] \\
& > \sE\bigl[\|\varepsilon_\mI\|^2\bigr],
\end{align}
which proves
\bea
\sE\bigl[\|\tZ_{\mI|t} - \tZ^*_\mI\|\bigr] < \sE\bigl[\|\tZ_\mI - \tZ^*_{\mI}\|\bigr]
\eea
The strict inequality holds whenever $\delta^*(t) \ne 0$.
\end{proof}

{\noindent \textbf{Discussion. }}
From this proof, we see that if the language prompt $t$ provides additional symmetry cues ($\delta^*(t)\neq 0$), then a FiLM layer can ``add back'' these missing cues into the visual features, reducing the overall error. Our choice $\gamma(z_t)=\mathbf{1}$ and $\beta(z_t)=\delta^*(t)$ is simply a constructive example illustrating FiLM's ability to perform an additive correction. In practice, $\gamma(\cdot)$ and $\beta(\cdot)$ are gradually learned to approximate this fix and achieve a lower error than relying on vision features alone.

\subsection{Why does SAPG work better?}\label{sec:sup_sapg_better}
\paragraph{Initialization.}

\begin{figure}[t]
    \centering
    \begin{minipage}[t]{.32\linewidth}
        \vspace{0pt}
        \centering
        \caption*{\scalebox{1.0}{GT}}
        {\includegraphics[width=\linewidth]{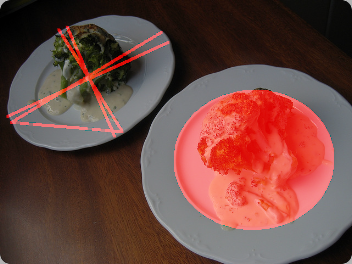}}
        {\includegraphics[width=\linewidth]{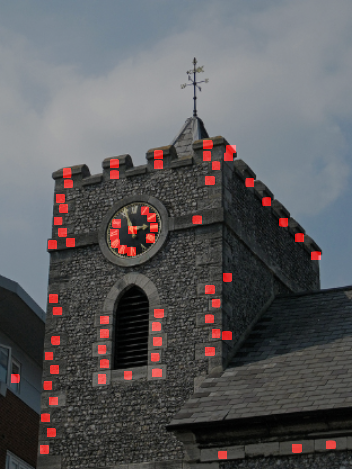}}
    \end{minipage}
    \begin{minipage}[t]{.32\linewidth}
        \vspace{0pt}
        \centering
        \caption*{\scalebox{1.0}{EquiSym~\citep{seo2022reflection}}}
        {\includegraphics[width=\linewidth]{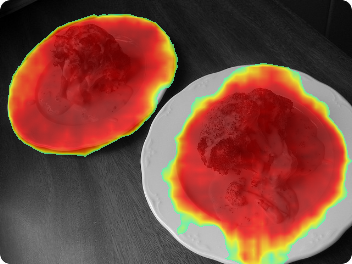}}
        {\includegraphics[width=\linewidth]{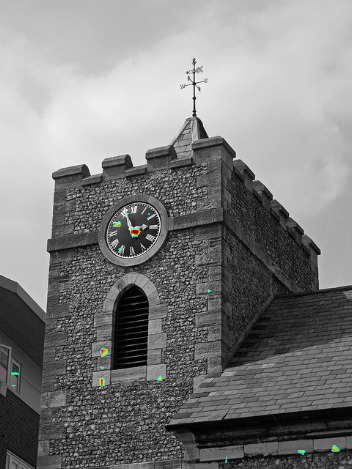}}
    \end{minipage}
    \begin{minipage}[t]{.32\linewidth}
        \vspace{0pt}
        \centering
        \caption*{\scalebox{1.0}{CLIPSym}}
        {\includegraphics[width=\linewidth]{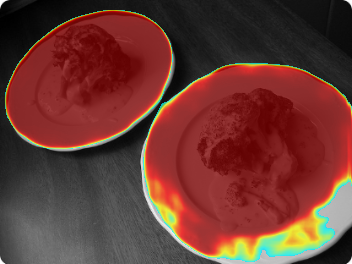}}
        {\includegraphics[width=\linewidth]{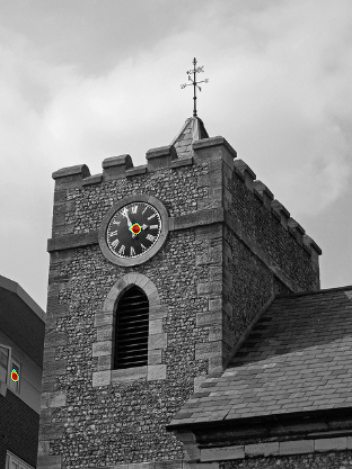}}
    \end{minipage}
    \caption{Examples of when symmeries cannot be well detected. The top row corresponds to the reflection case, and the bottom row corresponds to the rotation case.}
    \label{fig:vis_limit_sup}
\end{figure}

\begin{figure}[t]
    \centering
    \begin{minipage}[t]{.24\linewidth}
        \vspace{0pt}
        \centering
        \captionof*{figure}{GT}
        \vspace{-0.3cm}
        {\includegraphics[width=\linewidth]{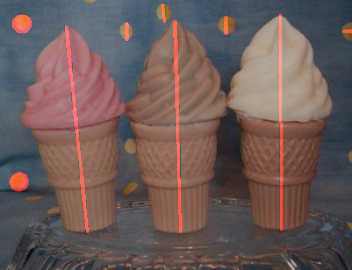}}
        \captionof*{figure}{GT}
        {\includegraphics[width=\linewidth]{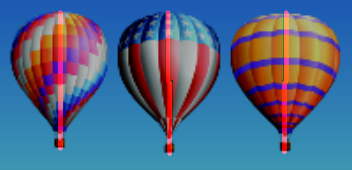}}
    \end{minipage}
    \begin{minipage}[t]{.24\linewidth}
        \vspace{0pt}
        \centering
        \captionof*{figure}{\emph{"cat"}}
        \vspace{-0.3cm}
        {\includegraphics[width=\linewidth]{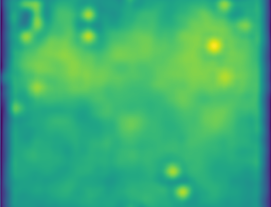}}
        \captionof*{figure}{\emph{"tree"}}
        {\includegraphics[width=\linewidth]{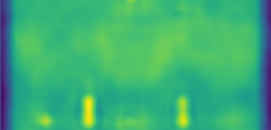}}
    \end{minipage}
    \begin{minipage}[t]{.24\linewidth}
        \vspace{0pt}
        \centering
        \captionof*{figure}{\emph{"ice cream"}}
        \vspace{-0.3cm}
        {\includegraphics[width=\linewidth]{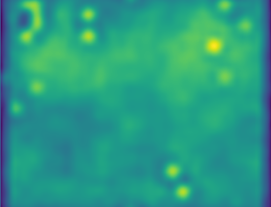}}
        \captionof*{figure}{\emph{"balloon"}}
        {\includegraphics[width=\linewidth]{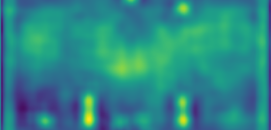}}
    \end{minipage}
    \begin{minipage}[t]{.24\linewidth}
        \vspace{0pt}
        \centering
        \captionof*{figure}{SAPG}
        \vspace{-0.3cm}
        {\includegraphics[width=\linewidth]{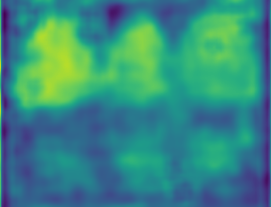}}
        \captionof*{figure}{SAPG}
        {\includegraphics[width=\linewidth]{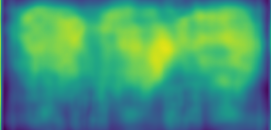}}
    \end{minipage}
    \caption{Illustrative examples of predicted symmetry heatmaps under different propmpting strategies at the initial step.}
    \label{fig:vis_sapg}
\end{figure}

Fig.~\ref{fig:vis_sapg} shows two illustrative examples of the predicted symmetry heatmaps under different prompting strategies at the initial step, i.e., before training. We can see that using unrelated random prompts causes the model to overly focus on most pixels in the image, while it fails to concentrate on regions that likely exhibit symmetry. For instance, in the second column, since \emph{"cat"} or \emph{"tree"} do not exist in the original images, the model cannot find the correct focus. In the third column, we design prompts specifically corresponding to symmetric objects, e.g., \emph{"ice cream"} and \emph{"balloon"}. Although the model's focus on symmetric objects improves compared to using unrelated random prompts, the distinction is still not enough. In the last column, the prompt aggregation via SAPG enables the model to correctly concentrate on symmetric objects. 

The underlying reason that this improved initialization is beneficial because it provides the model with a strong semantic prior derived from frequently occurring objects which are associated with symmetry cues. As a result, the model starts from a more informed state, reducing noise and misalignment in the early stages of training.

\paragraph{Prompt grouping. }
In contrast to a single prompt, which captures only one aspect of the semantic information and may suffer from high variance due to noisy or limited cues, aggregating multiple prompt-conditioned outputs via SAPG acts as an ensemble. While the predictions from individual prompts are indeed correlated due to the shared encoders, each prompt still focuses slightly different semantic cues about symmetry. Furthermore, since the aggregation weights are learnable so that the model can put more weights on prompts that are more aligned with symmetry. These factors lead to reduced noise and more stable predictions.

\paragraph{Why fixed prompts rather than adaptive ones?} Although one may consider using adaptive prompts that vary per image, we choose fixed prompts for several reasons. First, since CLIP's language encoder is trainable, as a result, the prompt embeddings gradually evolve during training to better capture the universal concept of symmetry. Second, adaptive prompts can be difficult to learn and may not generalize well. While it might be possible to identify an optimal prompt for each training image, generating new, adaptive prompt combinations for unseen images has the risk of producing semantic cues that do not reliably represent symmetry.

In the future, a possible direction is exploring image-correlated language embeddings, which has the potential to better represent more refined cues of symmetry based on each image's content.

\section{Additional results}\label{sec:sup_res}

\subsection{Limitations} 

Fig.~\ref{fig:vis_limit_sup} shows some cases in the DENDI dataset when symmetry cannot be well detected. The top row corresponds to the reflection case, while the bottom row corresponds to the rotation case.
The limitations are likely due to the dataset's annotation quality, such as inconsistency and ambiguity. For example, the left round object (plate) in the reflection example is not annotated as a circle as it usually should be, and the complicated rotation centers on the city wall in the rotation example are not obvious. We leave the improvement of the dataset quality for future work.

\subsection{More visualization results on DENDI dataset}
\begin{figure*}[t]
    \vspace{-0.1cm}
    \centering
    \begin{subfigure}[t]{0.49\textwidth}
    \begin{minipage}[t]{.24\linewidth}
    \vspace{0pt}
      \centering
      \caption*{\scalebox{1.0}{Ground Truth}}
        {\includegraphics[width=\linewidth]{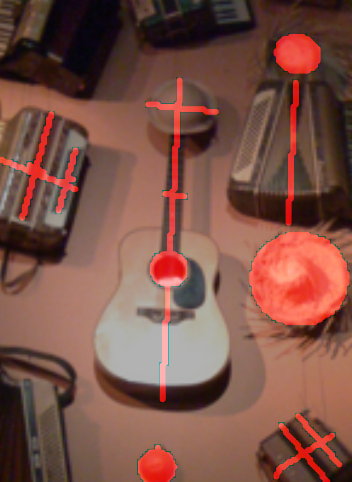}}
        {\includegraphics[width=\linewidth]{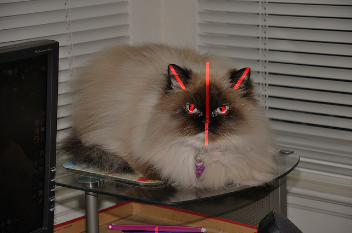}}
        {\includegraphics[width=\linewidth]{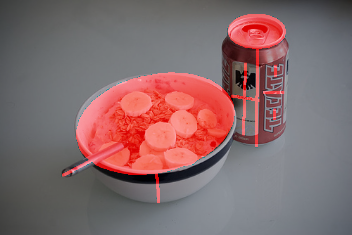}}
        {\includegraphics[width=\linewidth]{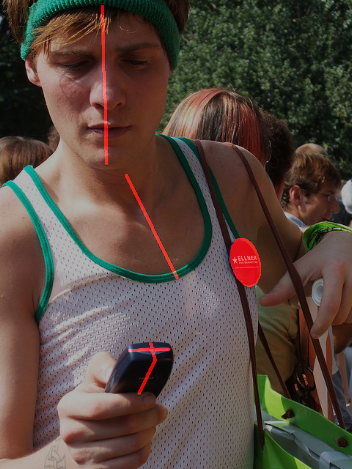}}
        {\includegraphics[width=\linewidth]{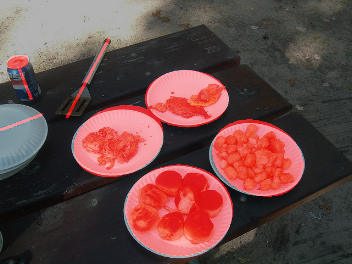}}
        {\includegraphics[width=\linewidth]{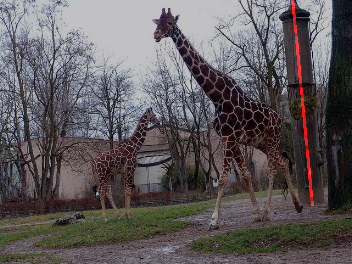}}
        {\includegraphics[width=\linewidth]{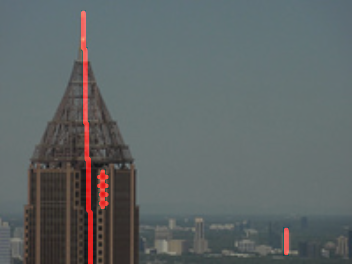}}
        {\includegraphics[width=\linewidth]{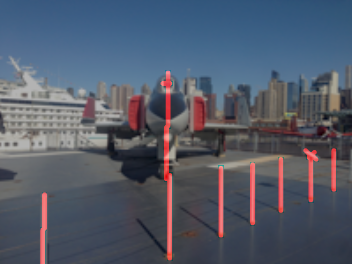}}
        {\includegraphics[width=\linewidth]{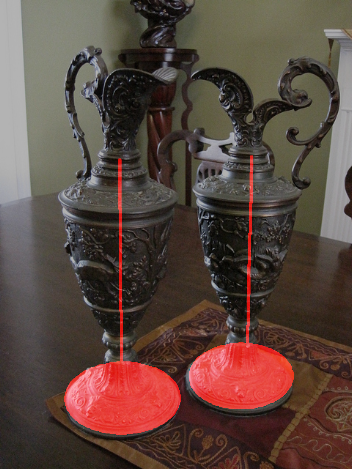}}
        {\includegraphics[width=\linewidth]{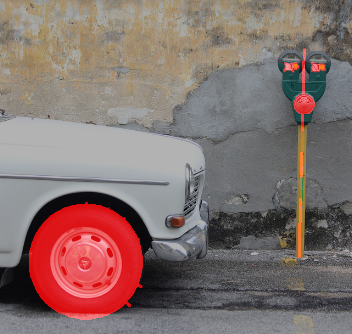}}
    \end{minipage}
    \begin{minipage}[t]{.24\linewidth}
    \vspace{0pt}
      \centering
      \caption*{\scalebox{1.0}{PMCNet~\citep{seo2021learning}}}
        {\includegraphics[width=\linewidth]{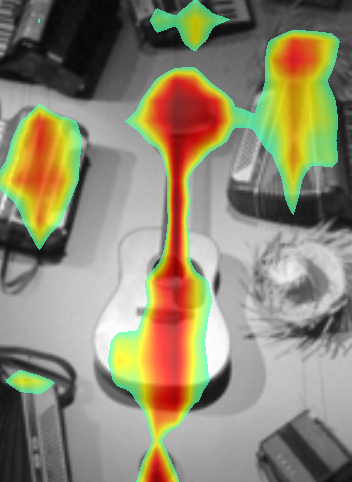}}
        {\includegraphics[width=\linewidth]{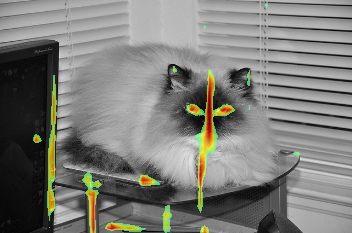}}
        {\includegraphics[width=\linewidth]{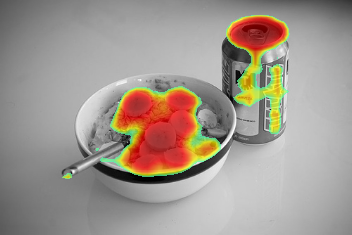}}
        {\includegraphics[width=\linewidth]{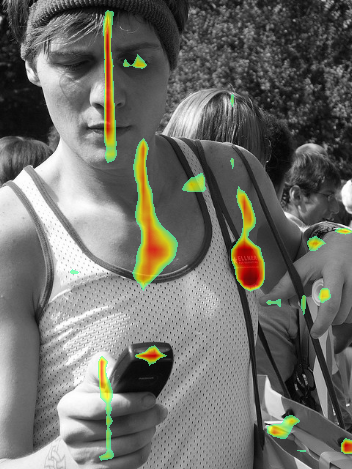}}
        {\includegraphics[width=\linewidth]{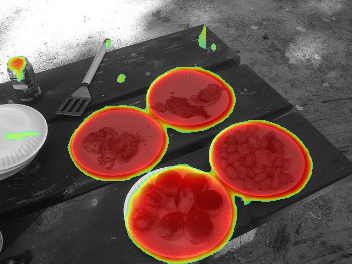}}
        {\includegraphics[width=\linewidth]{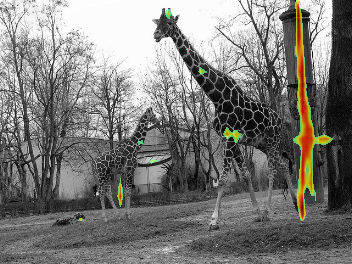}}
        {\includegraphics[width=\linewidth]{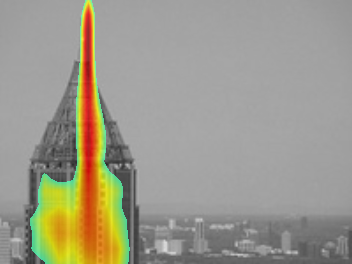}}
        {\includegraphics[width=\linewidth]{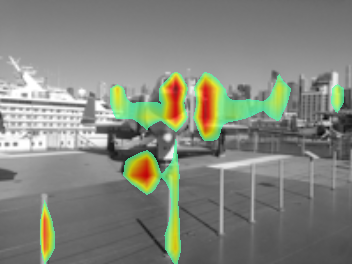}}
        {\includegraphics[width=\linewidth]{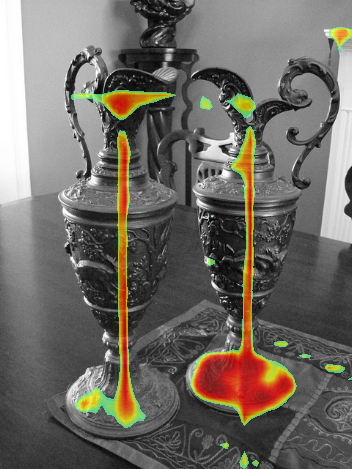}}
        {\includegraphics[width=\linewidth]{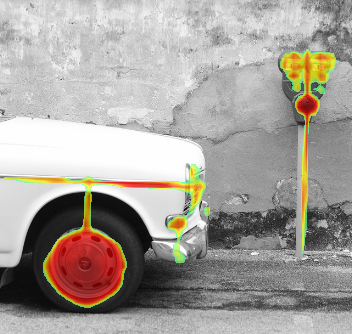}}
      \label{fig:vis_sup_ref_pmc}
    \end{minipage}
      \begin{minipage}[t]{.24\linewidth}
      \vspace{0pt}
      \centering
      \caption*{\scalebox{1.0}{EquiSym~\citep{seo2022reflection}}}
        {\includegraphics[width=\linewidth]{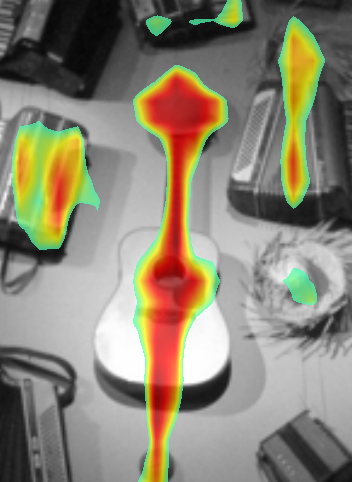}}
        {\includegraphics[width=\linewidth]{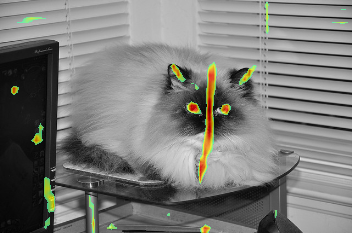}}
        {\includegraphics[width=\linewidth]{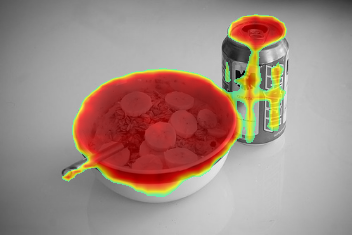}}
        {\includegraphics[width=\linewidth]{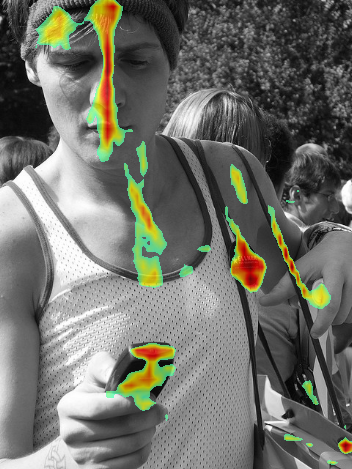}}
        {\includegraphics[width=\linewidth]{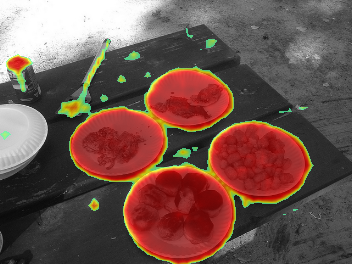}}
        {\includegraphics[width=\linewidth]{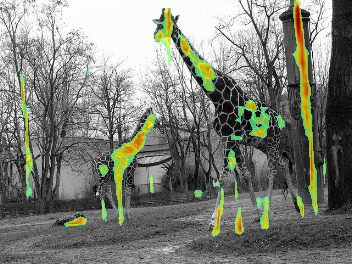}}
        {\includegraphics[width=\linewidth]{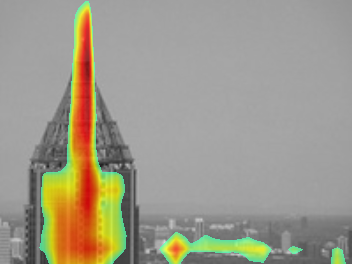}}
        {\includegraphics[width=\linewidth]{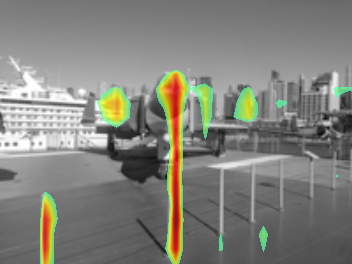}}
        {\includegraphics[width=\linewidth]{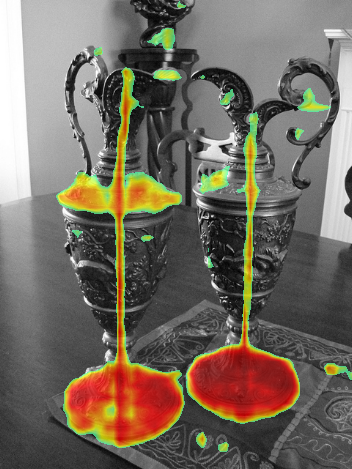}}
        {\includegraphics[width=\linewidth]{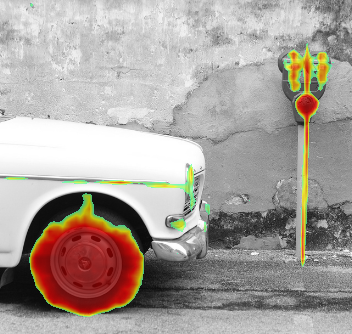}}
    \end{minipage}
        \begin{minipage}[t]{.24\linewidth}
        \vspace{0pt}
      \centering
      \caption*{\scalebox{1.0}{CLIPSym}}
        {\includegraphics[width=\linewidth]{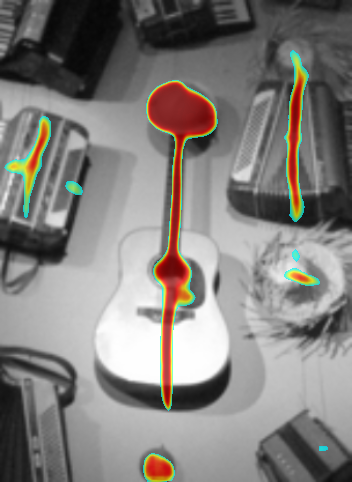}}
        {\includegraphics[width=\linewidth]{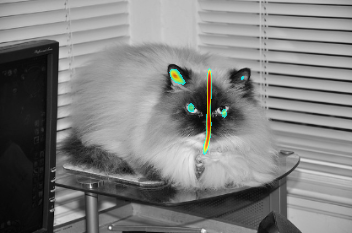}}
        {\includegraphics[width=\linewidth]{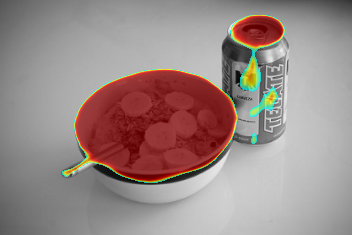}}
        {\includegraphics[width=\linewidth]{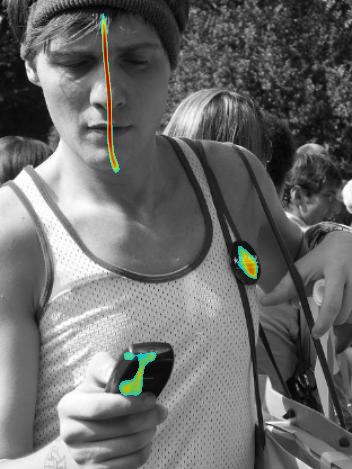}}
        {\includegraphics[width=\linewidth]{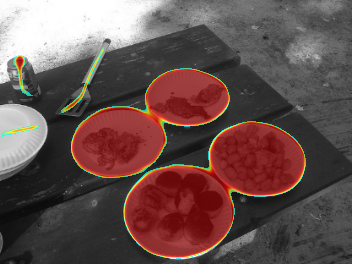}}
        {\includegraphics[width=\linewidth]{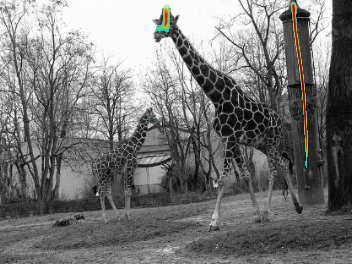}}
        {\includegraphics[width=\linewidth]{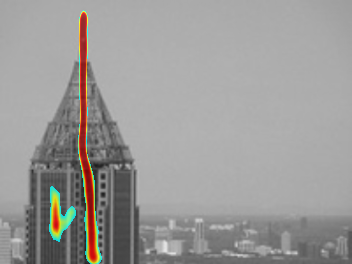}}
        {\includegraphics[width=\linewidth]{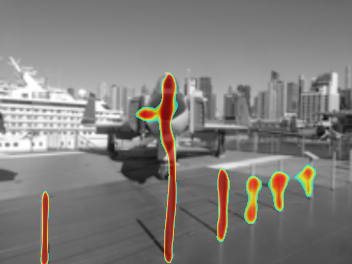}}
        {\includegraphics[width=\linewidth]{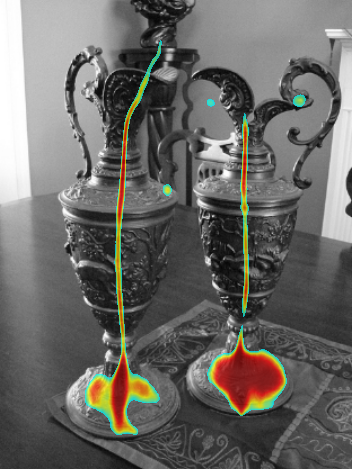}}
        {\includegraphics[width=\linewidth]{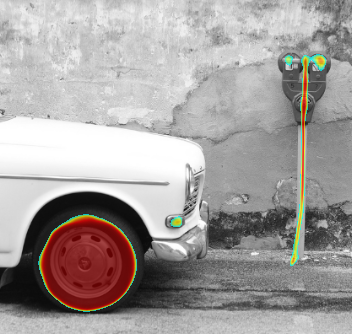}}
        \label{fig:vis_supp_ref_clipsym}
    \end{minipage}
    \vspace{-0.45cm}
  \caption{Reflection detection results on DENDI-\textit{ref}.}
  \label{fig:vis_supp_ref}
  \end{subfigure}
  \hspace{-0.05mm}
  \begin{subfigure}[t]{0.485\textwidth}
    \centering
    \begin{minipage}[t]{.32\linewidth}
    \vspace{0pt}
      \centering
      \caption*{\scalebox{1.0}{Ground Truth}}
        {\includegraphics[width=\linewidth]{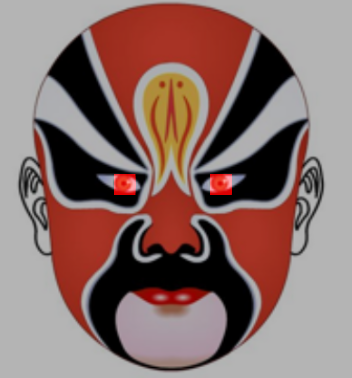}}
        {\includegraphics[width=\linewidth]{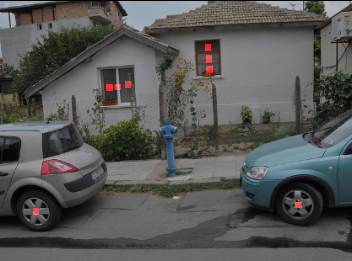}}
        {\includegraphics[width=\linewidth]{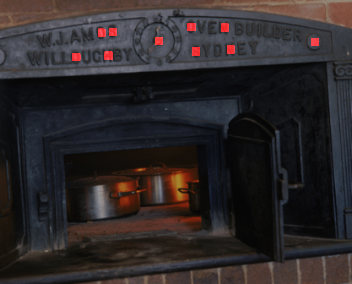}}
        {\includegraphics[width=\linewidth]{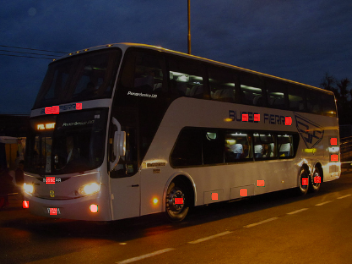}}
        {\includegraphics[width=\linewidth]{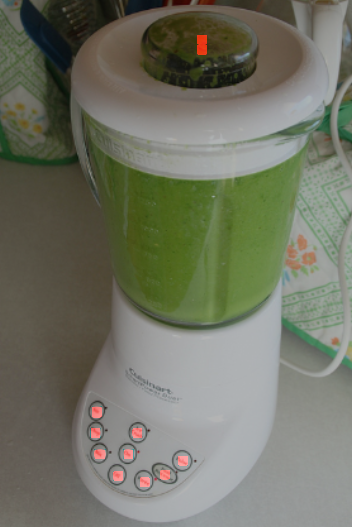}}
        {\includegraphics[width=\linewidth,trim={0 0 0 4.6cm},clip]{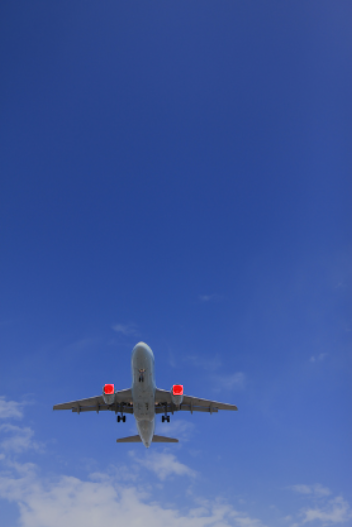}}
        {\includegraphics[width=\linewidth]{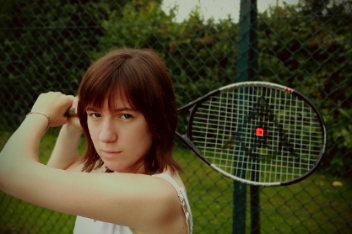}}
        {\includegraphics[width=\linewidth]{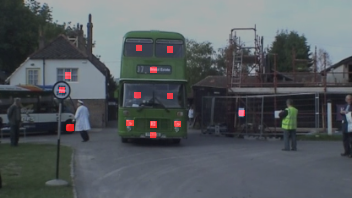}}
      
    \end{minipage}
    \begin{minipage}[t]{.32\linewidth}
    \vspace{0pt}
      \centering
      \caption*{\scalebox{1.0}{EquiSym~\citep{seo2022reflection}}}
        {\includegraphics[width=\linewidth]{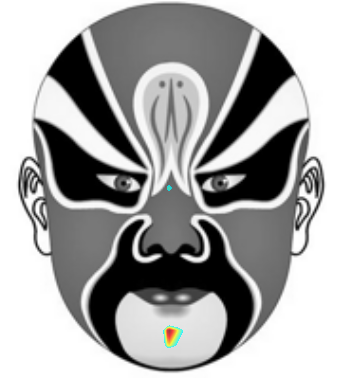}}
        {\includegraphics[width=\linewidth]{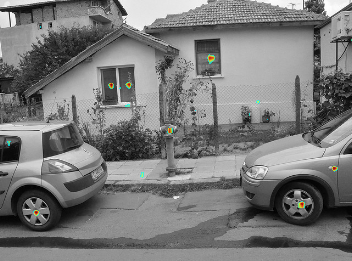}}
        {\includegraphics[width=\linewidth]{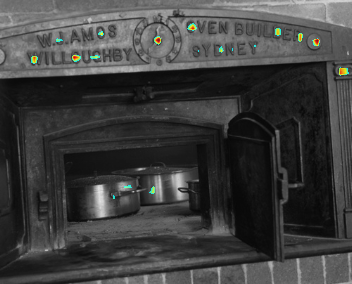}}
        {\includegraphics[width=\linewidth]{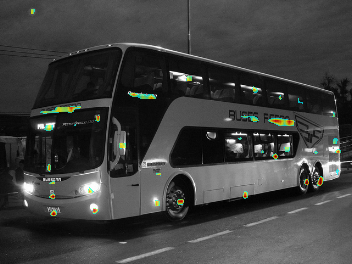}}
        {\includegraphics[width=\linewidth]{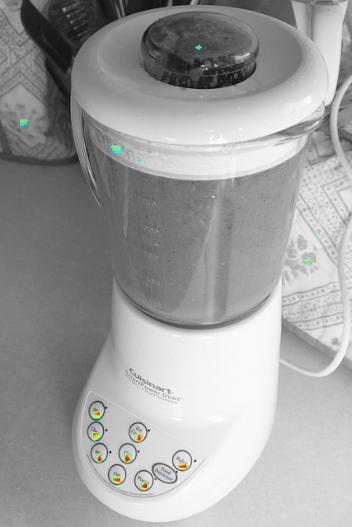}}
        {\includegraphics[width=\linewidth,trim={0 0 0 4.6cm},clip]{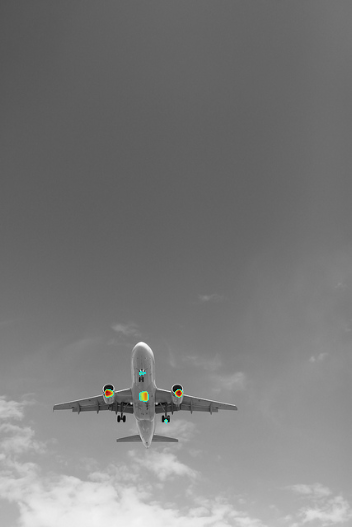}}
        {\includegraphics[width=\linewidth]{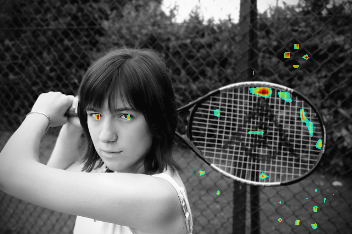}}
        {\includegraphics[width=\linewidth]{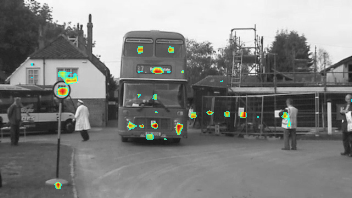}}
      \label{fig:rot_sup_eq}
    \end{minipage}
      \begin{minipage}[t]{.32\linewidth}
      \vspace{0pt}
      \centering
      \caption*{\scalebox{1.0}{CLIPSym}}
        {\includegraphics[width=\linewidth]{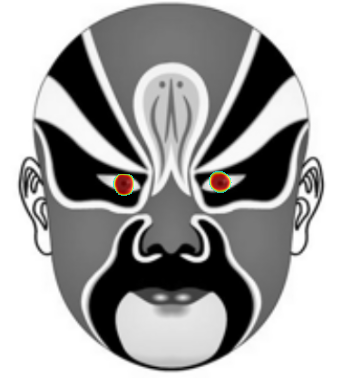}}
        {\includegraphics[width=\linewidth]{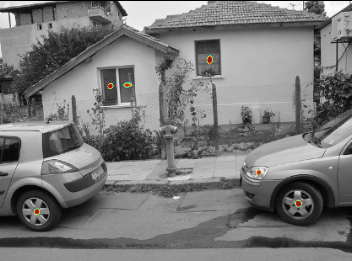}}
        {\includegraphics[width=\linewidth]{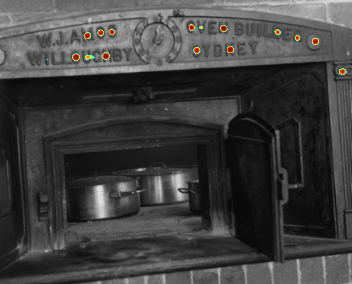}}
        {\includegraphics[width=\linewidth]{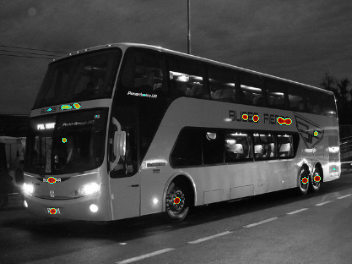}}
        {\includegraphics[width=\linewidth]{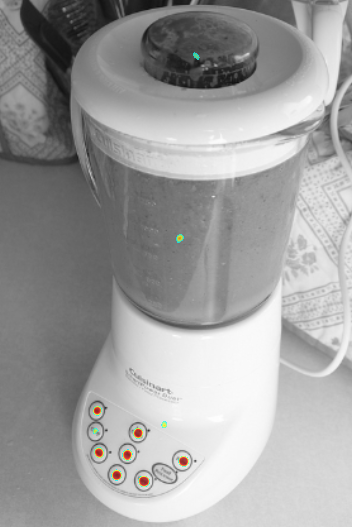}}
        {\includegraphics[width=\linewidth,trim={0 0 0 4.6cm},clip]{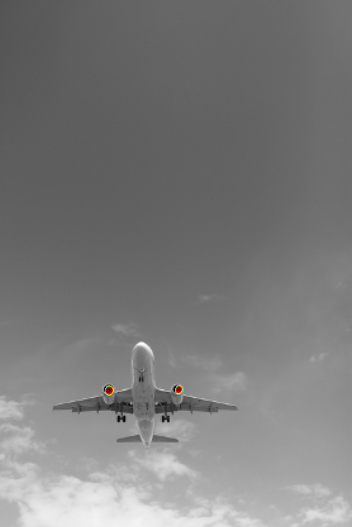}}
        {\includegraphics[width=\linewidth]{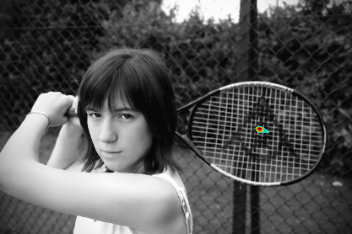}}
        {\includegraphics[width=\linewidth]{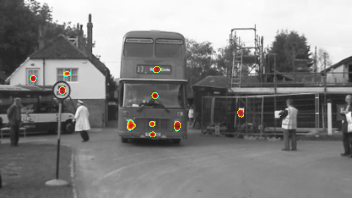}}
    \end{minipage}
    \vspace{-0.45cm}
    \caption{Rotation detection results on DENDI-\textit{rot}.}
    \label{fig:vis_supp_rot}
    \end{subfigure}
          \vspace{-0.1cm}
  \caption{Visualization of the reflection and rotation symmetry detection on the DENDI dataset.
  }
          \vspace{-0.1cm}
    \label{fig:vis_supp}
  \end{figure*}

Fig.~\ref{fig:vis_supp}
shows more qualitative results of reflection and rotation symmetry detection on the DENDI dataset. The results further show that CLIPSym can detect both reflection and rotation symmetries more effectively than other baselines.

\subsection{Consistency and robustness results on SDRW and LDRS datasets}\label{sub_sec:sup_cons_robust}

In Tab.~\ref{tab:con_rob_pmc}, we provide the consistency and robustness evaluation results for SDRW and LDRS reflection datasets under $[-45^\circ, 45^\circ]$ uniformly distributed rotation operations. Similar to results in Fig.~\ref{tab:con_rob}, CLIPSym achieves the best performance in terms of both robustness and consistency on every dataset that we evaluated.

\begin{table}[htb!]
    \setlength{\tabcolsep}{3pt}
    \centering
    \small
    \begin{tabular}{lcccccc}
    \specialrule{.15em}{.05em}{.05em}
    \multirow{2}{*}{\bf{Method}} & \multicolumn{2}{c}{\bf{SDRW}} & \multicolumn{2}{c}{\bf{LDRS}} & \multicolumn{2}{c}{\bf{Mixed}} \\
    & \bf{Rob.}${\bm \uparrow}$ & \bf{Cons.}${\bm \downarrow}$ & \bf{Rob.}${\bm \uparrow}$ & \bf{Cons.}${\bm \downarrow}$ & \bf{Rob.}${\bm \uparrow}$ & \bf{Cons.}${\bm \downarrow}$ \\
    \hline
    PMCNet~\citep{seo2021learning} & 40.4 & 0.263 & 21.6 & 0.356 & 25.0 & 0.333 \\
    EquiSym~\citep{seo2022reflection} & 39.4 & 0.101 & 20.5 & 0.112 & 24.8 & 0.109 \\ %
    CLIPSym$^\text{non-eq.}$ & 42.2 & 0.059 & 27.3 & 0.061 & 30.2 & 0.060 \\
    CLIPSym$^\text{eq.}$ & {\bf 44.3} & {\bf 0.042} & {\bf 29.2} & {\bf 0.042} & {\bf 32.3} & {\bf 0.042} \\
    \specialrule{.15em}{.05em}{.05em}
    \end{tabular}
    \vspace{-0.15cm}
    \caption{Equivariance robustness and consistency evaluation results for SDRW and LDRS reflection datasets under $[-45^\circ, 45^\circ]$ uniformly distributed rotation operations. 
    }
    \vspace{-0.1cm}
    \label{tab:con_rob_pmc}
    \end{table}

\subsection{Visualizations of robustness and consistency}
\begin{figure*}[t]
    \vspace{-0.1cm}
    \centering
    \begin{subfigure}[t]{1.0\textwidth}
    \begin{minipage}[t]{.137\linewidth}
        \vspace{0pt}
        \centering
        \caption*{\scalebox{0.75}{Original image}}
        {\includegraphics[width=\linewidth]{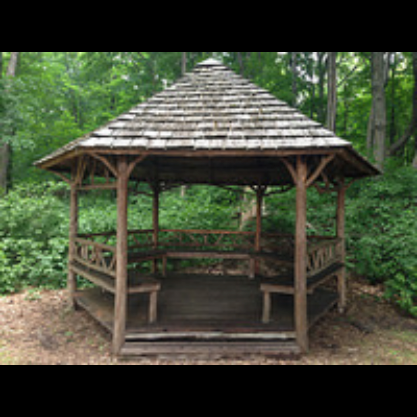}}
        {\includegraphics[width=\linewidth]{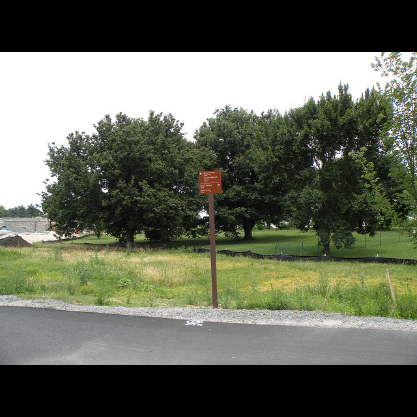}}
        {\includegraphics[width=\linewidth]{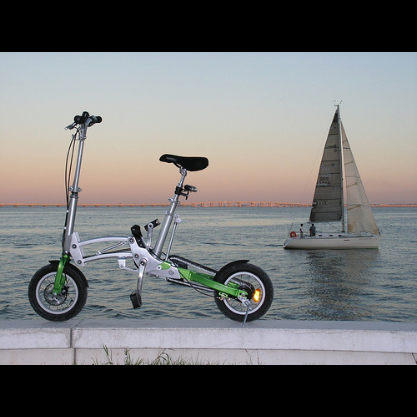}}
        {\includegraphics[width=\linewidth]{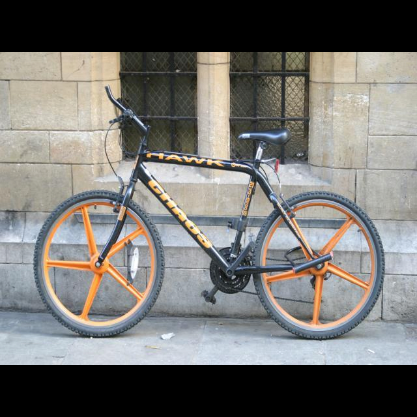}}
        {\includegraphics[width=\linewidth]{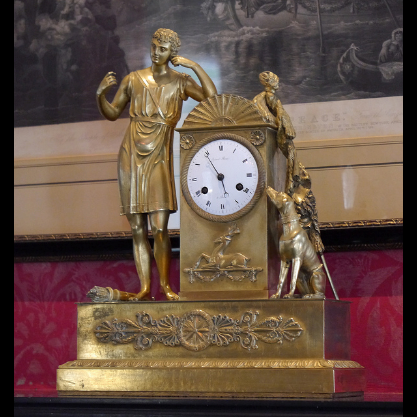}}
        {\includegraphics[width=\linewidth]{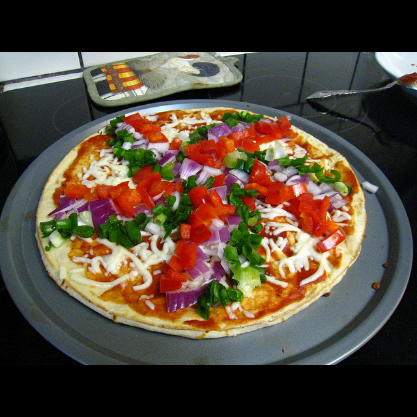}}
        {\includegraphics[width=\linewidth]{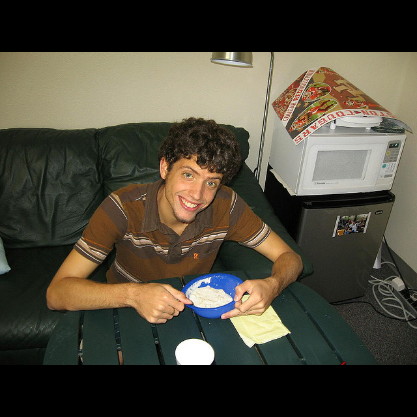}}
        {\includegraphics[width=\linewidth]{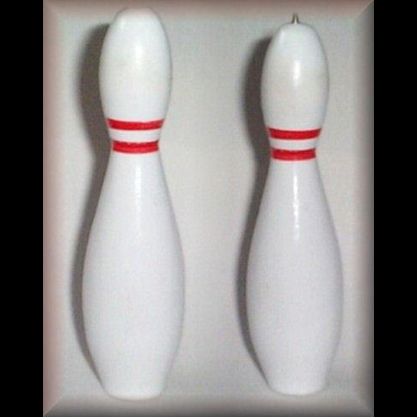}}
    \end{minipage}
    \begin{minipage}[t]{.137\linewidth}
        \vspace{0pt}
        \centering
        \caption*{\scalebox{0.75}{GT}}
        {\includegraphics[width=\linewidth]{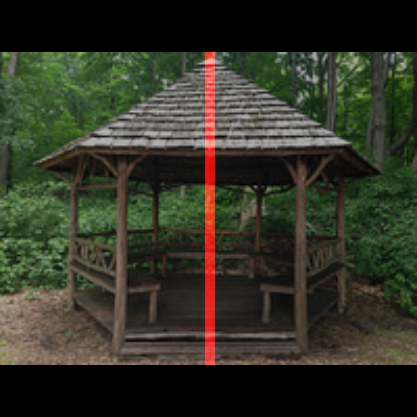}}
        {\includegraphics[width=\linewidth]{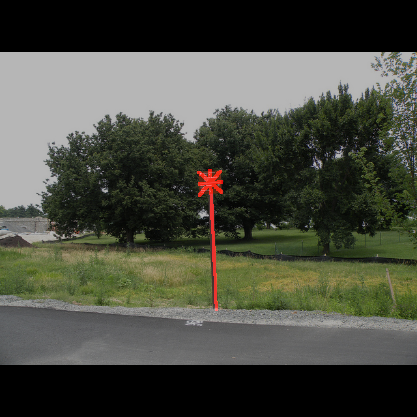}}
        {\includegraphics[width=\linewidth]{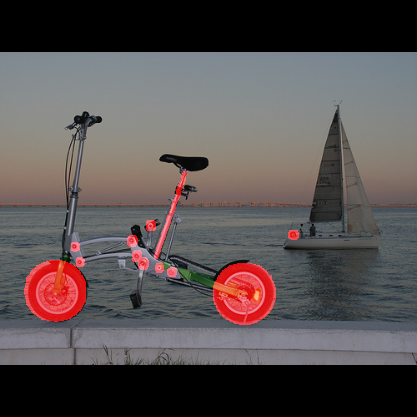}}
        {\includegraphics[width=\linewidth]{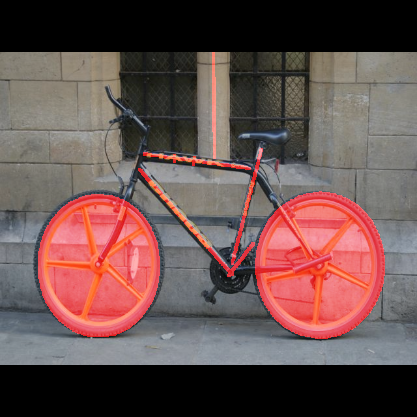}}
        {\includegraphics[width=\linewidth]{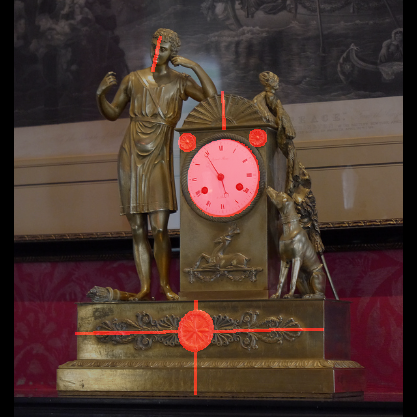}}
        {\includegraphics[width=\linewidth]{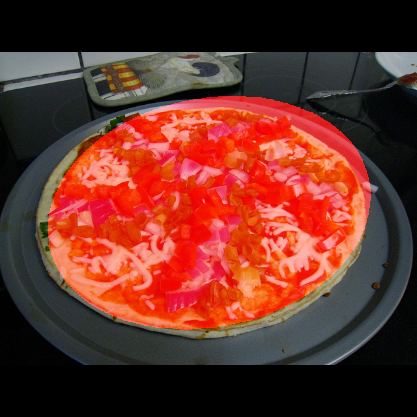}}
        {\includegraphics[width=\linewidth]{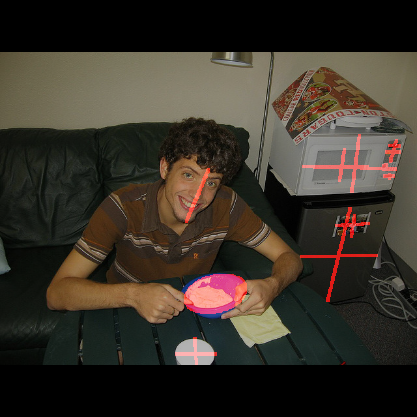}}
        {\includegraphics[width=\linewidth]{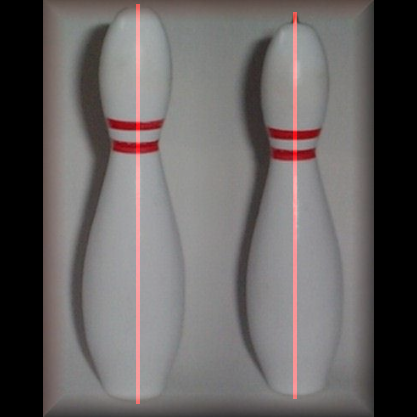}}
    \end{minipage}
    \begin{minipage}[t]{.137\linewidth}
        \vspace{0pt}
        \centering
        \caption*{\scalebox{0.75}{Rotated GT}}
        {\includegraphics[width=\linewidth]{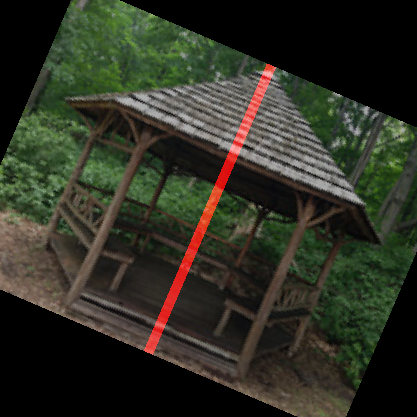}}
        {\includegraphics[width=\linewidth]{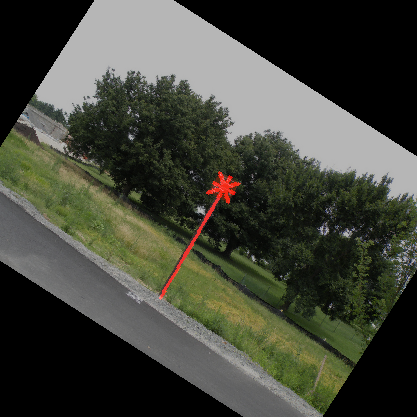}}
        {\includegraphics[width=\linewidth]{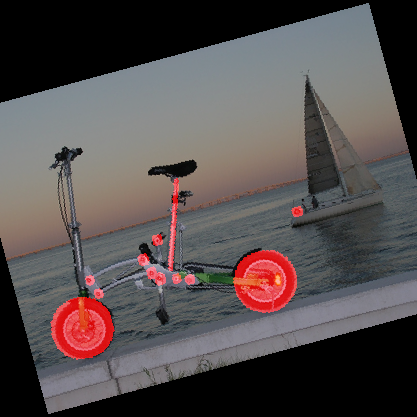}}
        {\includegraphics[width=\linewidth]{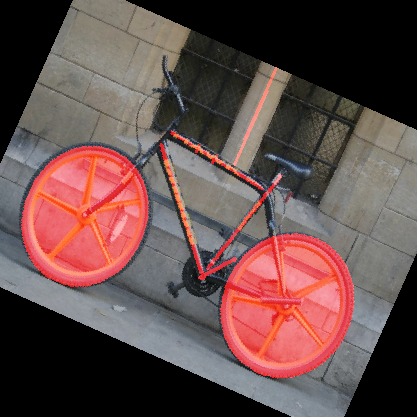}}
        {\includegraphics[width=\linewidth]{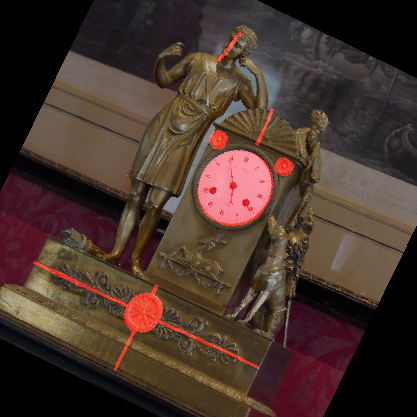}}
        {\includegraphics[width=\linewidth]{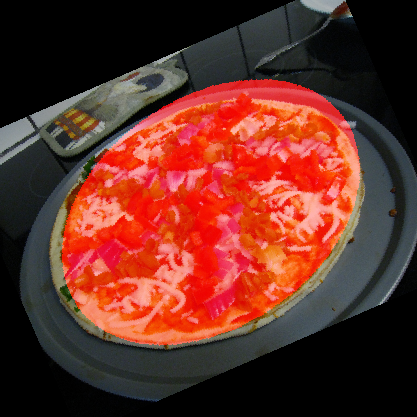}}
        {\includegraphics[width=\linewidth]{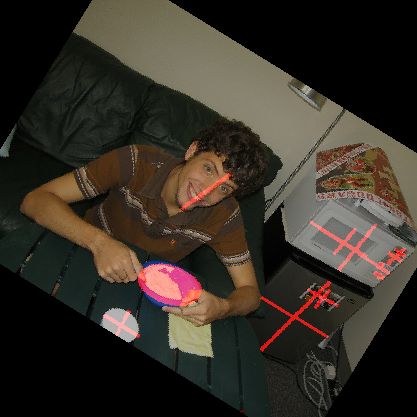}}
        {\includegraphics[width=\linewidth]{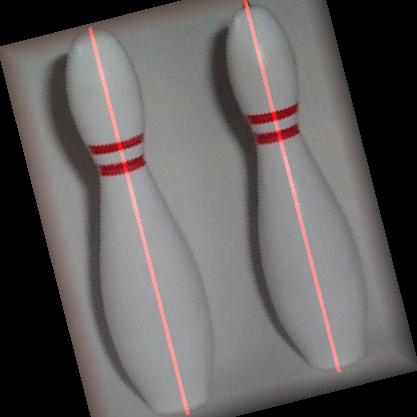}}
    \end{minipage}
    \begin{minipage}[t]{.137\linewidth}
        \vspace{0pt}
        \centering
        \caption*{\scalebox{0.75}{EquiSym $\hat{\mS}_{\mT(\mI)}$}}
        {\includegraphics[width=\linewidth]{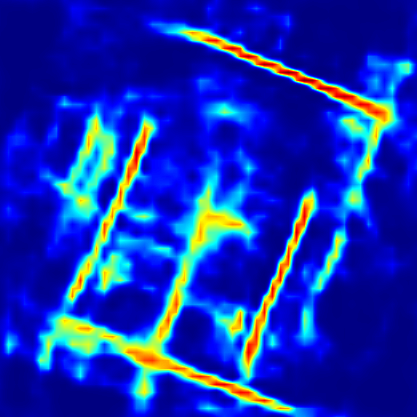}}
        {\includegraphics[width=\linewidth]{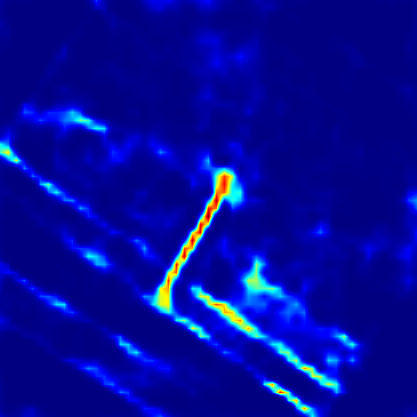}}
        {\includegraphics[width=\linewidth]{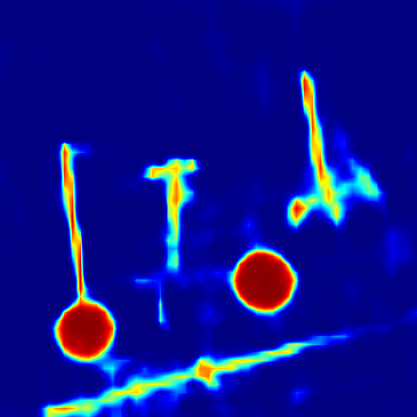}}
        {\includegraphics[width=\linewidth]{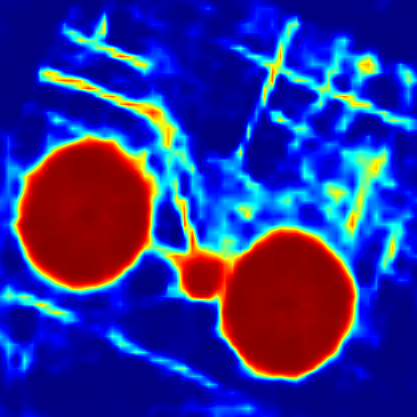}}
        {\includegraphics[width=\linewidth]{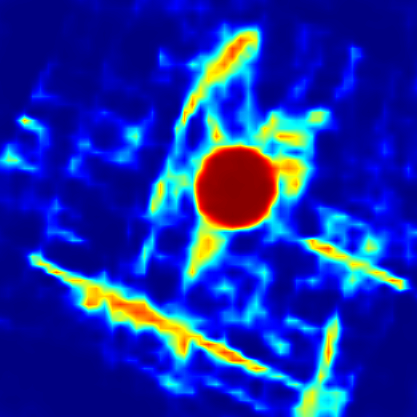}}
        {\includegraphics[width=\linewidth]{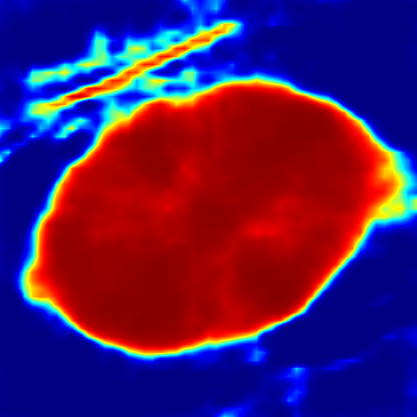}}
        {\includegraphics[width=\linewidth]{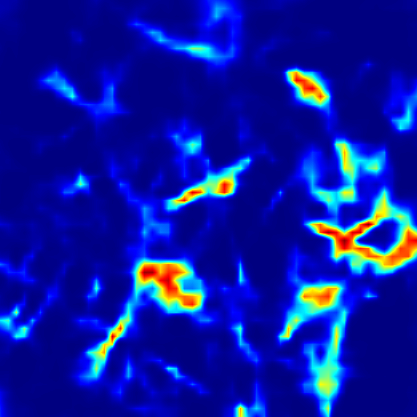}}
        {\includegraphics[width=\linewidth]{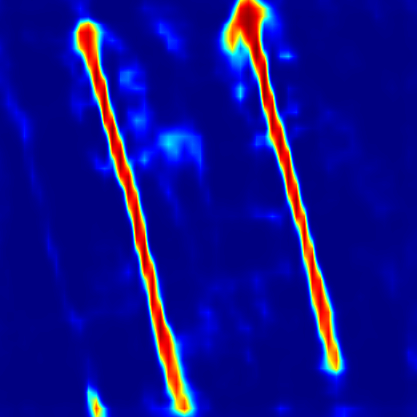}}
    \end{minipage}
    \begin{minipage}[t]{.137\linewidth}
        \vspace{0pt}
        \centering
        \caption*{\scalebox{0.75}{EquiSym $\mT(\hat{\mS}_\mI)$}}
        {\includegraphics[width=\linewidth]{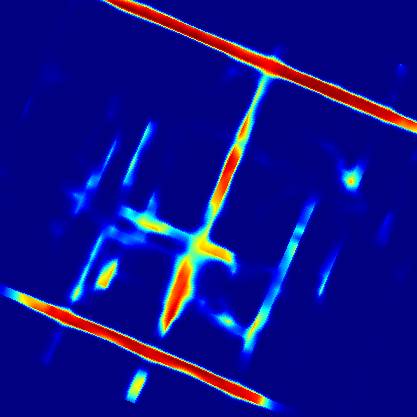}}
        {\includegraphics[width=\linewidth]{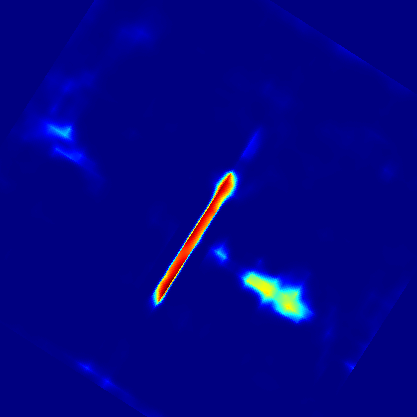}}
        {\includegraphics[width=\linewidth]{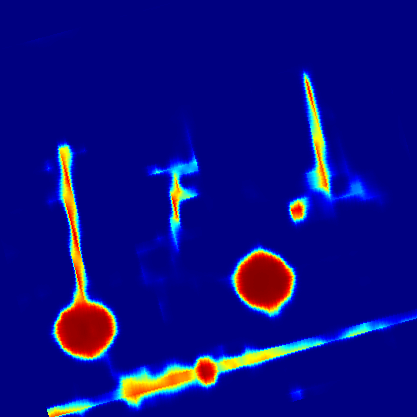}}
        {\includegraphics[width=\linewidth]{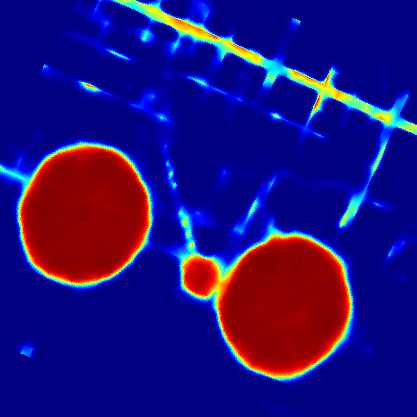}}
        {\includegraphics[width=\linewidth]{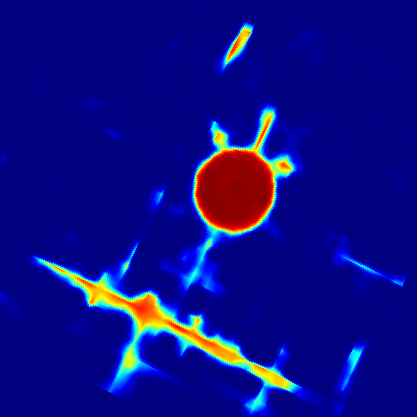}}
        {\includegraphics[width=\linewidth]{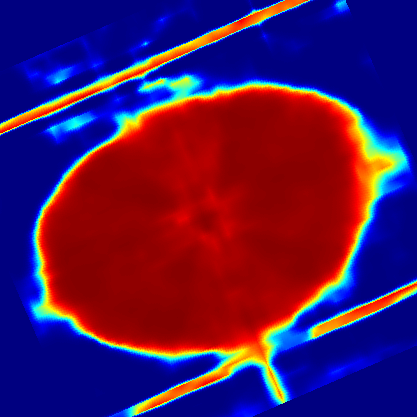}}
        {\includegraphics[width=\linewidth]{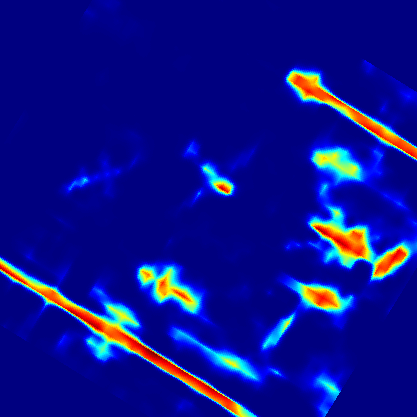}}
        {\includegraphics[width=\linewidth]{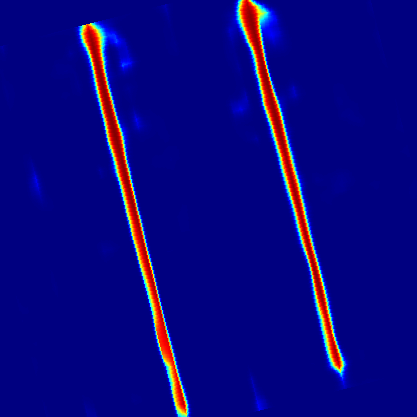}}
    \end{minipage}
    \begin{minipage}[t]{.137\linewidth}
        \vspace{0pt}
        \centering
        \caption*{\scalebox{0.75}{CLIPSym $\hat{\mS}_{\mT(\mI)}$}}
        {\includegraphics[width=\linewidth]{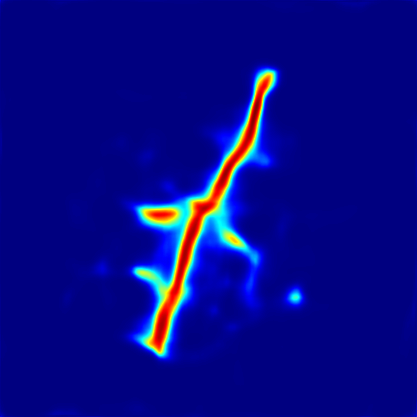}}
        {\includegraphics[width=\linewidth]{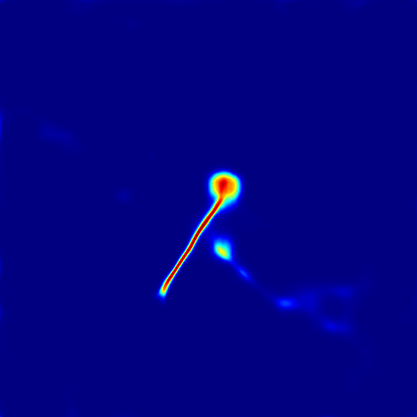}}
        {\includegraphics[width=\linewidth]{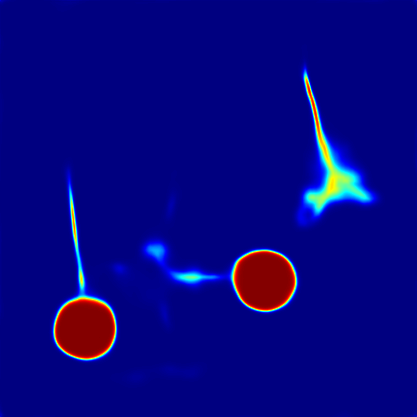}}
        {\includegraphics[width=\linewidth]{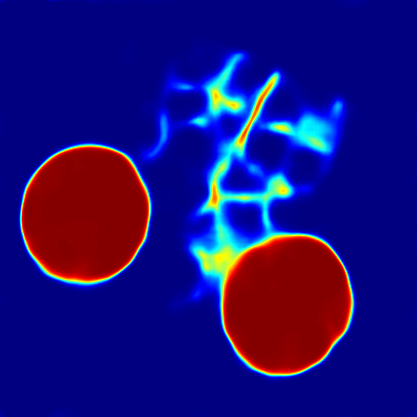}}
        {\includegraphics[width=\linewidth]{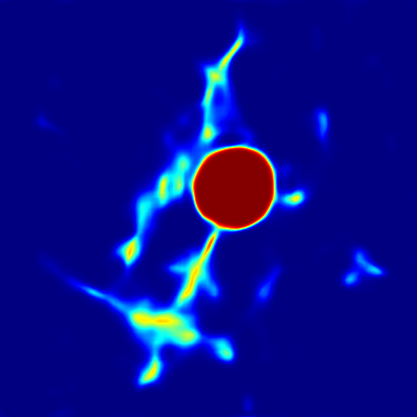}}
        {\includegraphics[width=\linewidth]{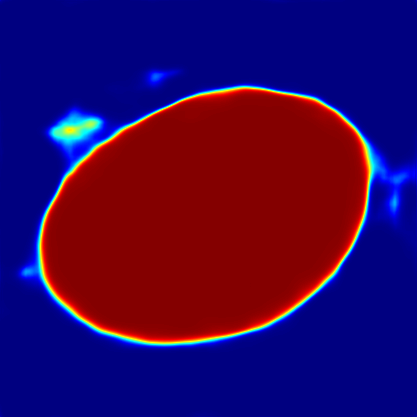}}
        {\includegraphics[width=\linewidth]{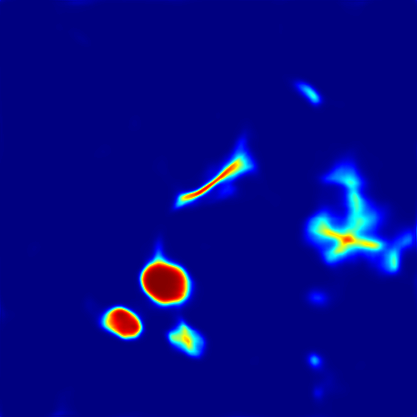}}
        {\includegraphics[width=\linewidth]{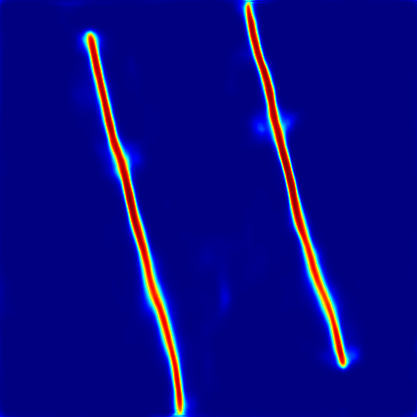}}
    \end{minipage}
    \begin{minipage}[t]{.137\linewidth}
        \vspace{0pt}
        \centering
        \caption*{\scalebox{0.75}{CLIPSym $\mT(\hat{\mS}_\mI)$}}
        {\includegraphics[width=\linewidth]{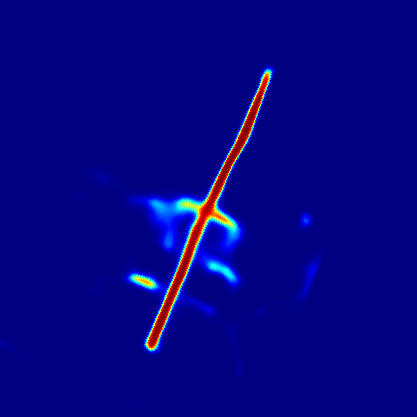}}
        {\includegraphics[width=\linewidth]{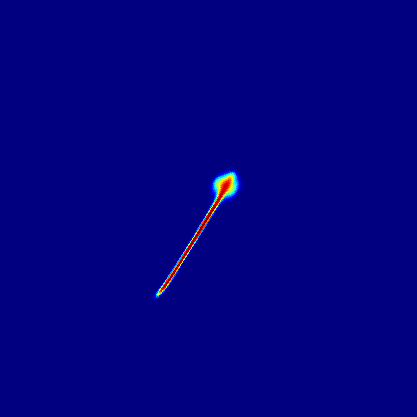}}
        {\includegraphics[width=\linewidth]{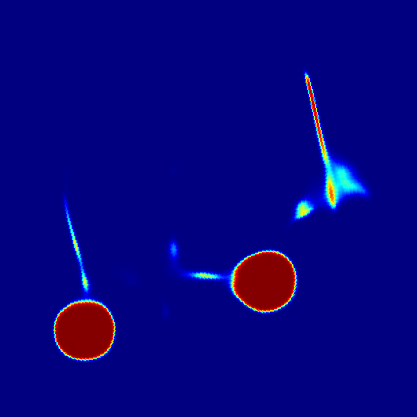}}
        {\includegraphics[width=\linewidth]{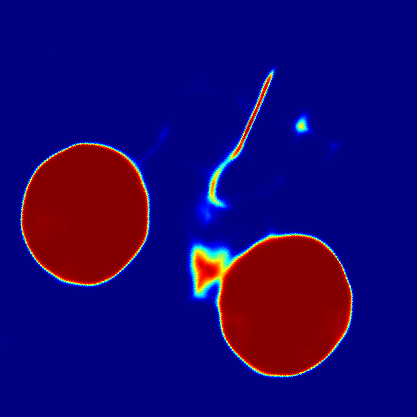}}
        {\includegraphics[width=\linewidth]{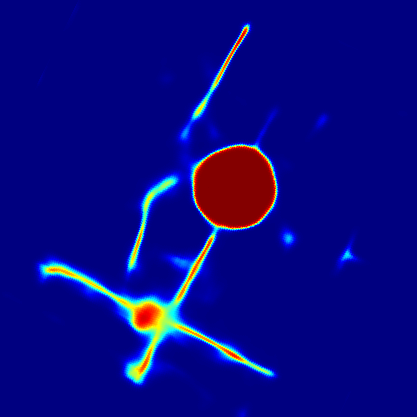}}
        {\includegraphics[width=\linewidth]{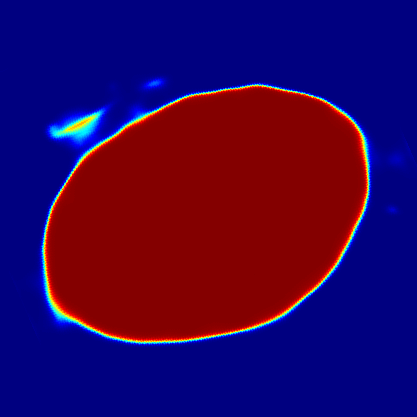}}
        {\includegraphics[width=\linewidth]{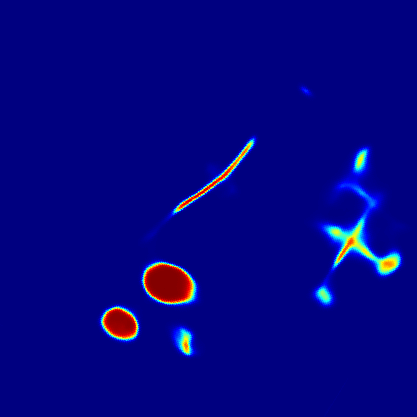}}
        {\includegraphics[width=\linewidth]{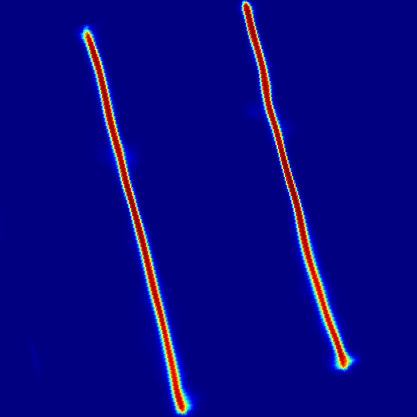}}
    \end{minipage}
  
  \end{subfigure}
  \caption{Examples of original image, ground truth, rotated ground truth, EquiSym and CLIPSym's predicted heatmaps $\hat{\mS}_{\mT(\mI)}$ on the rotated image and the rotated heatmap $\mT(\hat{\mS}_\mI)$.
  }
  \label{fig:sup_vis_rot}
  \vspace{-0.6cm}
  \end{figure*}

In~\figref{fig:cons_vis_rot}, we present heatmaps of EquiSym and CLIPSym which take images under random rotation transformations within $[-45^\circ, 45^\circ]$ as inputs to illustrate the model's robustness and consistency. In~\figref{fig:sup_vis_rot}, we present more results in addition to~\figref{fig:cons_vis_rot}.

\subsection{Ablation study on different equivariance degrees}
In~\tabref{tab:sup_group_eq}, we evaluate the F1-score on DENDI reflection dataset under different degrees of group-equivariance in the design of CLIPSym decoder. Specifically, we evaluate $C_4$-, $C_6$-, $C_8$-, $C_{12}$- and $C_{16}$-equivariant decoders. The results show that $C_8$-equivariant decoder, which is the same as our model's setting, achieves the best performance.

\setlength{\tabcolsep}{6pt}
\begin{table}[htb!]
    \centering
    \caption{Quantitative comparison of F1-score (\%) on the DENDI~\cite{seo2022reflection} reflection dataset under different degrees of group-equivariance.
    }
    \vspace{2mm}
    \label{tab:sup_group_eq}
    \begin{tabular}{cccccc}
        \toprule
        {\bf Equiv. degrees} & $C_4$ & $C_6$ & $C_8$ & $C_{12}$ & $C_{16}$ \\
        \midrule
        {\bf Ref. F1} & 65.3 & 64.3 & {\bf 66.5}  &  65.6 & 65.8 \\
        \bottomrule
    \end{tabular}
\end{table}

\subsection{The best prompts structure for symmetry detection}
We conduct hyperparameter search over the number of prompts $M \in \{1, 10, 25, 50\}$ and the number of object classes in each prompt $K \in \{1, 4, 8, 16, 32\}$, as well as generating different combinations of objects using different seeds, the best prompts structure in our experiments of both reflection and rotation symmetry detection has $M=25$ and $K=4$ and is given as \tabref{tab:best_prompts}.
\begin{table*}[ht]
    \setlength{\tabcolsep}{7pt}
    \caption{Details of the best prompts structure for symmetry detection.}
    \centering
    \begin{tabular}{c|ccccc}
        \toprule
        & $t_1$ & $t_2$ & $t_3$ & $t_4$ & $t_5$ \\
        \midrule
        {\tt obj}$_1$ & tundra & ski pole & alphabet & tennis racket & race track \\
        {\tt obj}$_2$ & orangutan & pass & snowboard & bright & martini glass \\
        {\tt obj}$_3$ & maze & take & cap & control & windshield \\
        {\tt obj}$_4$ & antler & sibling & champagne & construction site & stone building \\
        \midrule
        & $t_6$ & $t_7$ & $t_8$ & $t_9$ & $t_{10}$ \\
        \midrule
        {\tt obj}$_1$ & damage & breakfast & sedan & pot & meter \\
        {\tt obj}$_2$ & mouth organ & liquor & ceramic & pasture & skater \\
        {\tt obj}$_3$ & driver & ladle & roll & driftwood & tripod \\
        {\tt obj}$_4$ & snout & motorboat & toilet seat & folding chair & monster \\
        \midrule
        & $t_{11}$ & $t_{12}$ & $t_{13}$ & $t_{14}$ & $t_{15}$ \\
        \midrule
        {\tt obj}$_1$ & coaster & padlock & mansion & bike lane & polka dot \\
        {\tt obj}$_2$ & ceremony & concrete & scale & brownie & out \\
        {\tt obj}$_3$ & shower door & spaghetti & zombie & color & keyboard \\
        {\tt obj}$_4$ & signature & bee & potato & wine bottle & autumn \\
        \midrule
        & $t_{16}$ & $t_{17}$ & $t_{18}$ & $t_{19}$ & $t_{20}$ \\
        \midrule
        {\tt obj}$_1$ & food stand & mill & spinach & crack & package \\
        {\tt obj}$_2$ & design & handbag & underwater & teal & coffee \\
        {\tt obj}$_3$ & scone & wet & knot & level & side table \\
        {\tt obj}$_4$ & camouflage & clothe & deliver & tape & coconut \\
        \midrule
        & $t_{21}$ & $t_{22}$ & $t_{23}$ & $t_{24}$ & $t_{25}$ \\
        \midrule
        {\tt obj}$_1$ & dart & urban & enclosure & pottery & buffet \\
        {\tt obj}$_2$ & city street & sweatshirt & screen door & bookcase & gravel \\
        {\tt obj}$_3$ & blackberry & bowl & turret & shelter & apple \\
        {\tt obj}$_4$ & goalkeeper & bike lane & tricycle & squad & rearview mirror \\
        \bottomrule
    \end{tabular}
    \vspace{0.2cm}
    \label{tab:best_prompts}
\end{table*}

\subsection{Variant of non-equivariant version of CLIPSym with 8x decoder channels}
The only difference between CLIPSym's the non-equivariant version (CLIPSym$^\text{non-eq.}$) and the equivariant version CLIPSym$^\text{eq.}$ is whether the upsampler is equivariant or not. In CLIPSym$^\text{non-eq.}$ we use regular CNN, while in CLIPSym$^\text{eq.}$ we use $G$-convolution. Both models have the same number of channel dimensions $[64, 32, 16, 1]$ and $3 \times 3$ filters. To ensure a fair comparison to highlight the design of the equivariant module indeed helps, we also evaluate CLIPSym$^\text{non-eq.}$ with $8$ times more channels in~\tabref{tab:8times_no_eq}. Even with the similar decoder capacity, by comparing with results in Tab.~\ref{tab:dendi} and Tab.~\ref{tab:pmc_test}, CLIPSym$^\text{eq.}$ outperforms this larger baseline, demonstrating the benefits of the equivariant design.
\begin{table}[H]
  \centering
  \small
  \vspace{-8pt}
  \resizebox{\columnwidth}{!}{%
  \begin{tabular}{lccccc}
    \toprule
    Dataset & DENDI Ref. & DENDI Rot. & SDRW & LDRS & Mixed\\
    \midrule
    F1 & 64.9 $\pm$ 0.2 & 24.4 $\pm$ 0.1 &49.5 $\pm$ 0.3 & 37.6 $\pm$ 0.2 & 41.4 $\pm$ 0.2\\
    \bottomrule
  \end{tabular}
  }
  \vspace{-0.3cm}
  \caption{CLIPSym$^\text{non-eq.}$ with $8$ times more channels.}
  \label{tab:8times_no_eq}
  \vspace{-13pt}
\end{table}

\subsection{Comparison with an emerging result}
\citet{seo2025leveraging} recently proposed to lift 2-D features into the camera’s 3-D space and regress a seed vertex, which leads to a higher F1 score for rotation center detection on DENDI. However, their pipeline is restricted to rotation symmetry, which requires an extra 2D to 3D label conversion, and cannot handle reflection. In contrast, \textbf{CLIPSym} offers a \emph{unified} solution for both reflection \textit{and} rotation.

\section{Additional implementation details}\label{sec:sup_imp}

{\bf \noindent Image processing.} To conduct a fair comparison with other baselines, images with different resolutions are reshaped to $417 \times 417$ pixels as in~\citep{seo2022reflection} and~\citep{seo2021learning} before feeding into the model. We preserve the aspect ratio of the original images and pad the shorter side with zeros. 
During training, data augmentations include random rotations of intervals of 90 degrees, random small rotations within $[-15^\circ, 15^\circ]$, and color jittering.
Since the image encoder uses ViT-B/16~\citep{dosovitskiy2020image}, the size of each image patch is $16 \times 16$ pixels, and the number of patches of each side of images is $M = \lfloor 417 / 16 \rfloor = 26$. During testing, we still follow the same image processing pipeline as in training but without data augmentations to fit the model's input size. When calculating the metrics, zero paddings are cropped and images are resized back to their original sizes.

{\bf \noindent Decoder details. }
The decoder's FiLM block described in~\secref{subsec:decoder} modulates the image features conditioned on text prompts. To conduct the element-wise multiplication in~\equref{eqn:flim}, we project the text token features and image token features to the same dimension using linear layers with learnable parameters. The dimension of the linear layer is set to $d = 64$. The Transformer module consists of $L_{\etB} = 3$ multi-headed attention layers, while the upsampler contains 3$\times$ $G$-conv. and bi-linear interpolation submodules.

{\bf \noindent Focal loss.} In the $\alpha$-focal loss defined in~\equref{eq:focal_loss}, we have
\begin{align}
    & \hat{\mS}'_{\mI_{xy}} =  \left\{ \begin{array}{l}
          \hat{\mS}_{\mI_{xy}} ~~ \text{if } \mS_{\mI_{xy}} = 1 \\
         1 - \hat{\mS}_{\mI_{xy}} ~~ \text{otherwise}
    \end{array} \right. \\
        & \alpha'_{\mI_{xy}} =  \left\{ \begin{array}{l}
          \alpha ~~ \text{if } \mS_{\mI_{xy}} = 1 \\
         1 - \alpha ~~ \text{otherwise.}
    \end{array} \right.
\end{align}
where $\alpha$ represents the balance factor between the symmetry and non-symmetry pixels. 
We set $\alpha=0.85$ for reflection symmetry detection and $\alpha=0.95$ for rotation since reflection axes contain more positive pixels than rotation centers.
The focusing parameter $\lambda$ in \equref{eq:focal_loss} is set to 2.0.

{\bf \noindent Other training details.} We trained CLIPSym for 500 epochs with a batch size of 16 on a single NVIDIA A100 80GB GPU. A single epoch takes around 5 minutes and the total training takes approximately 40 hours. 
We conducted a hyperparameter search within $\{10^{-3}, 10^{-4}, 5\times 10^{-5}, 10^{-5}, 5\times 10^{-6}, 10^{-6}\}$ and found $10^{-5}$ is the best initial learning rate for both reflection and rotation symmetry detection. The difference is that reflection detection uses an exponential decay scheduler with a decay rate of 0.1, while rotation detection uses a constant learning rate.

\end{document}